\documentclass[runningheads]{llncs}
\usepackage{graphicx}
\usepackage{hyperref}
\usepackage{booktabs}
\usepackage{float}
\usepackage{amsmath}
\usepackage{multirow}
\usepackage{xcolor}
\usepackage{tabularx}
\usepackage{array}

\newcolumntype{Y}{>{\raggedright\arraybackslash}X}
\newcolumntype{L}[1]{>{\raggedright\arraybackslash}p{#1}}

\begin{document}

\title{
The Algorithmic Caricature: Auditing LLM-Generated Political Discourse Across Crisis Events
}
\titlerunning{The Algorithmic Caricature}

\author{Gunjan, Sidahmed Benabderrahmane, Talal Rahwan}
\authorrunning{Gunjan et al.}
\institute{New York University (NYUAD), Division of Science, Computer Science Department\\
\email{gg2713@nyu.edu}}

\maketitle

\begin{abstract}
Large Language Models (LLMs) are increasingly capable of generating fluent political text, raising concerns about the large-scale production of synthetic discourse during periods of social conflict. While prior detection work has often emphasized local linguistic cues such as perplexity, burstiness, or token-level irregularities, these signals may become less reliable as generative systems improve. In this paper, we adopt a Computational Social Science (CSS) perspective and study synthetic political discourse at the population level rather than the sentence level. 

We construct a paired multi-event corpus of 1,789,406 posts spanning nine politically salient crisis events: the COVID-19 pandemic, the Jan 6 Capitol attack, the 2020 US election, the 2024 US election, Dobbs/Roe v. Wade, the 2020 BLM protests, the US midterm elections, the Utah shooting, and the US-Iran war. For each event, we compare \textit{observed} online discourse collected from social platforms with \textit{synthetic} discourse generated to discuss the same event. We evaluate divergence along four dimensions: emotional intensity, structural regularity, lexical-ideological framing, and cross-event dependency, using mean-based gaps together with dispersion-based evidence of population distortion. 

Across events, synthetic discourse is more negative and less dispersed in sentiment, structurally more regular, and lexically more abstract than observed discourse. In contrast, observed discourse exhibits broader emotional dispersion, longer-tailed structural distributions, and more context-specific and colloquial lexical markers. These differences are not uniform across events: they are typically larger for fast-moving, decentralized crises and smaller for formal, institutionally mediated events. We summarize these differences through a simple event-level divergence measure, which we term the \textit{Caricature Gap}. 

Our findings suggest that the main limitation of synthetic political discourse is not grammatical fluency, but reduced population realism. We argue that this population-level perspective offers a useful complement to traditional text-detection approaches and provides a CSS framework for auditing the social realism of generated discourse.

\textbf{Data availability:} Code and data are available at \url{https://github.com/gunjan2713/AuditingSyntheticPoliticalLanguage}
\end{abstract}

\section{Introduction}

Social media platforms have become central infrastructures for political expression, mobilization, and contestation \cite{hasan2025role}. They are also key sites of manipulation, including coordinated influence operations, misinformation campaigns, and automated amplification \cite{orgeret2026hashtags},\cite{ahmed2026youth}. The rapid diffusion of Large Language Models (LLMs) raises a new concern within this landscape: the possibility of generating large volumes of fluent, politically charged synthetic discourse that can imitate grassroots expression at scale \cite{lu2026llm}.

Much of the recent literature on AI-generated text detection has focused on local textual signatures, including token predictability, burstiness, repetition, or perplexity-based irregularities \cite{kehkashan2025ai}. Such methods can be useful, but they are increasingly vulnerable to model improvements, paraphrasing, and stylistic adaptation. In settings such as political communication, where language is noisy, emotional, and event-dependent, a narrow focus on sentence-level cues may miss broader population-level distortions \cite{wolfsfeld2022making}.

In this paper, we adopt a Computational Social Science (CSS) perspective and ask a different question: \textit{when LLMs are used to simulate political discourse around crisis events, how does the resulting discourse differ from observed online populations?} Rather than treating synthetic text solely as an authorship problem, we treat it as a socio-linguistic population-level phenomenon. The contribution of this paper is not a new detector of AI-generated text, but a computational social science framework for auditing whether synthetic discourse populations reproduce the aggregate behavioral signatures of observed online publics. This distinction is important because political influence does not operate only through isolated texts; it also operates through population-level distributions of emotion, toxicity, repetition, lexical framing, and event-specific variability. By treating the discourse population rather than the individual post as the unit of analysis, the proposed framework connects generative AI evaluation to central CSS concerns: collective behavior, political communication, crisis response, and the measurement of online publics.

To investigate this question, we build a paired multi-event corpus containing both observed and synthetic discourse across nine crisis events. We then compare these two sources of discourse along four dimensions: emotional intensity, structural regularity, lexical-ideological framing, and cross-event dependency. Across these dimensions, we find consistent differences between observed and synthetic discourse. Synthetic text is typically more emotionally concentrated, structurally more homogeneous, and lexically more abstract or dramatized than observed discourse. At the same time, the magnitude of these divergences depends on the event type: differences tend to widen for decentralized and fast-moving crises and narrow for more institutional or formalized events.

Our contributions are fourfold:
\begin{enumerate}
    \item We introduce a paired multi-event benchmark for comparing observed and synthetic political discourse across nine crisis events.
    \item We propose a four-pillar CSS framework for auditing synthetic discourse at the population level: emotional intensity, structural regularity, lexical-ideological framing, and cross-event dependency.
    \item We formalize an event-level sociological typology that treats crisis events as different discourse environments rather than interchangeable topics.
    \item We show that synthetic discourse differs systematically from observed discourse in ways that are socially meaningful and event-dependent, and we summarize these differences through a simple divergence measure that we term the \textit{Caricature Gap}.
\end{enumerate}

Overall, the paper argues for a shift in emphasis: from asking whether a single text looks machine-generated to asking whether a generated discourse population behaves like a real one.

\subsection{Research Questions and Hypotheses}

This study is guided by three research questions:

\textbf{RQ1.} Do LLM-generated political discourse populations differ from observed online discourse populations across crisis events?

\textbf{RQ2.} Which dimensions of discourse realism are most affected: emotional intensity, structural regularity, lexical-ideological framing, or cross-event dependency?

\textbf{RQ3.} Are synthetic--observed divergences stable across events, or are they moderated by the sociological character of the event?

Based on prior work on synthetic social data, political communication, and machine-generated text, we formulate three descriptive hypotheses. First, we expect synthetic discourse to exhibit lower population heterogeneity than observed discourse, reflected in narrower sentiment, length, and lexical distributions. Second, we expect divergence to be larger for fast-moving and decentralized crises, where observed discourse is more informal and heterogeneous. Third, we expect toxicity divergence to be event-dependent rather than uniform, because model safety behavior and prompt framing may interact with the dominant rhetorical register of each event.

\section{Related Work}
This paper intersects three lines of research: automated detection of machine-generated text, computational social science studies of online political discourse, and the use of large language models to simulate or synthesize social data. We review each in turn and identify a gap that motivates a population-level treatment of crisis-era discourse.

The detection of machine-generated text has emerged as a central response to the rapid proliferation of large language models, with prior surveys organizing the field into statistical, stylometric, supervised, perturbation-based, and watermarking approaches \cite{crothers2023machine},\cite{wu2025survey}. Early statistical methods exploited the tendency of generated text to occupy high-likelihood, low-rank regions of a model's probability distribution, as exemplified by GLTR \cite{gehrmann2019gltr}, while later perturbation-based detectors such as DetectGPT \cite{mitchell2023detectgpt} and Binoculars \cite{hans2024binoculars} formalized this intuition through curvature and cross-perplexity criteria, achieving strong zero-shot performance. Supervised classifiers built on transformer backbones such as RoBERTa have likewise demonstrated high in-domain accuracy, with systems like Grover \cite{zellers2019grover} and Ghostbuster \cite{verma2024ghostbuster} separating human- from machine-written content with near-perfect F1 scores. Watermarking schemes embed pseudorandom signals at generation time to enable post-hoc verification \cite{kirchenbauer2023watermark}, though they presuppose cooperation from the generator and offer little protection against adversaries who
deploy unwatermarked open-source models. Stylometric techniques tailored to short-form content, particularly tweets, complement these methods by leveraging phraseology, punctuation, and lexical-diversity features \cite{fagni2021tweepfake}, \cite{kumarage2023stylometric}. Yet recent stress tests have shown that recursive paraphrasing and adversarial perturbations sharply degrade the reliability of all four families \cite{sadasivan2023can}, \cite{dugan2024raid}, revealing that contemporary detectors operate primarily at the local textual level and remain blind to the broader discursive and social patterns through which generated content circulates and exerts influence in conflict-driven online environments.

The second line of work comes from Computational Social Science and the study of online political discourse \cite{theocharis2024introduction}. This literature has examined polarization, affective language, misinformation, ideological clustering, protest communication, and event-driven shifts in online behavior. Our work builds on this tradition by analyzing discourse as a population-level social signal rather than merely a collection of isolated texts. Cross-platform analyses have established that online political interactions cluster into homophilic communities whose structure shapes information diffusion and exposure to opposing views \cite{cinelli2021echo}, and multinational evidence has further shown that out-group interactions are consistently more toxic than in-group ones, a regularity now treated as a defining signature of affective polarization \cite{falkenberg2024patterns}. This measurement tradition has been extended through methods that operationalize affective polarization via the emotion and toxicity of reply interactions, demonstrating that ideological proximity in the network correlates strongly with the valence and civility of expressed language \cite{he2024affective}. Crisis-driven events have provided especially rich settings for such analysis, with longitudinal studies tracking partisan dynamics during the COVID-19 pandemic \cite{chen2020covid}, the January 6th Capitol attack \cite{lee2022storm}, the overturning of Roe v. Wade \cite{chang2023roe}, and the 2024 U.S. presidential election \cite{balasubramanian2024public}, revealing sharp emotional, lexical, and ideological asymmetries between opposing user communities. Yet across this body of work, analyses are conducted on observed human-authored content alone, leaving open how the same population-level regularities behave when discourse is no longer
exclusively human in origin.

The third line of work concerns synthetic social data and simulation. Recent studies have explored the use of LLMs to emulate individuals, generate survey responses, or approximate social interactions. Our work differs by focusing specifically on crisis-driven political discourse and by asking whether generated discourse reproduces the emotional, structural, and lexical variability of observed online populations. Early efforts in this direction established the notion of \textit{algorithmic fidelity}, demonstrating that LLMs conditioned on socio-demographic backstories can reproduce response distributions of real survey participants across political subgroups
\cite{argyle2023outofone}. This insight has since been extended to richer settings, including sandbox environments populated by generative agents that plan, converse, and form social ties \cite{park2023generative}, and simulated micro-blogging platforms in which politically representative LLM populations are used to evaluate feed-ranking interventions and cross-partisan toxicity \cite{tornberg2023simulating}. In parallel, a growing body of evidence has raised concerns about the behavioral and ideological properties of synthetic content: LLM-generated tweets are at least as credible as human ones for both accurate information and disinformation \cite{spitale2023gpt3}, commercial models exhibit measurable ideological skews inherited from pretraining \cite{feng2023pretraining}, and large-scale agent-based opinion simulations converge toward artificial consensus that fails to reproduce the polarization observed in real online communities \cite{chuang2024simulating}. Despite these advances, evaluation has remained tethered either to individual-level fidelity to survey responses or to abstract network-level dynamics, leaving the question of whether synthetic discourse reproduces the population-level statistical signatures of authentic crisis-era human communication
largely unanswered.

Taken together, these three lines of research leave a gap: there is no systematic, multi-event, population-level CSS audit of whether synthetic political discourse reproduces the behavioral signatures of real online populations during crisis events. This paper fills that gap by introducing a paired corpus benchmark, a four-pillar analytical framework, and a simple descriptive divergence measure---the \textit{Caricature Gap}---that can be applied to any new event or generation system.

\section{Data and Paired Event Design}

\subsection{Overview}

Our dataset contains 1,789,406 posts distributed across nine crisis events: the COVID-19 pandemic, the Jan 6 Capitol attack, the 2020 US election, the 2024 US election, Dobbs/Roe v. Wade, the 2020 BLM protests, the US midterm elections, the Utah shooting, and the recent US-Iran war. For each event, we assemble two aligned corpora:
\begin{itemize}
    \item \textbf{Observed}: posts collected from public online platforms, including Twitter, Telegram, and Reddit.
    \item \textbf{Synthetic}: posts generated to discuss the same event using LLM-based generation pipelines.
\end{itemize}

We use the term \textit{observed} rather than \textit{human} because public social media may include bots, coordinated messaging, or machine-assisted content. Our contrast is therefore between \textit{observed online discourse} and \textit{synthetic generated discourse}.
\begin{figure}[h]
    \centering
    \includegraphics[width=\textwidth]{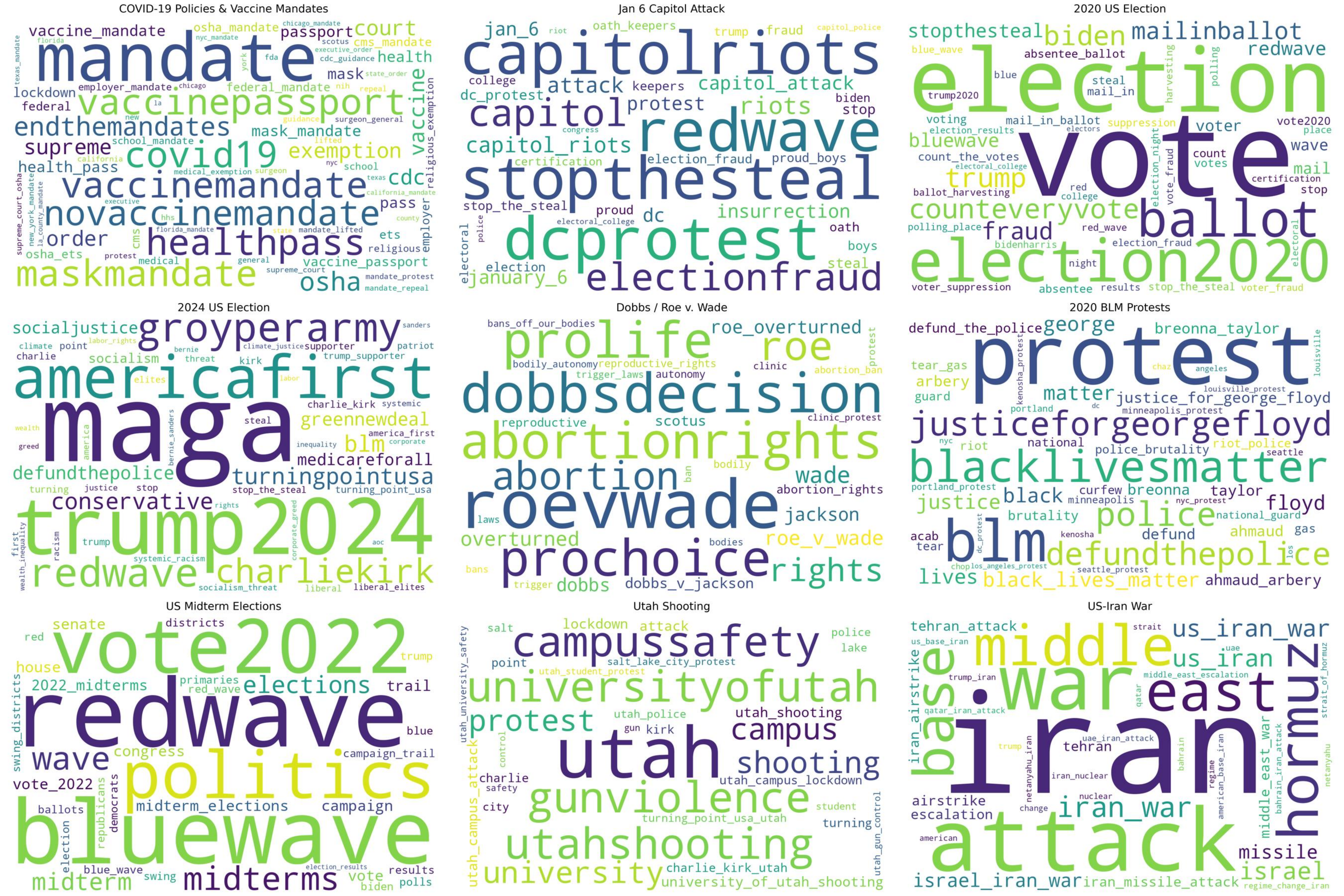}
    \caption{Event-specific scraping seed lexicons. Word clouds summarize the hashtags and keywords used to seed data collection for each crisis event. These visualizations describe the collection strategy rather than empirical word frequencies in the collected corpus.}
    \label{fig:seed-wordclouds}
\end{figure}

Figure~\ref{fig:seed-wordclouds} provides a visual summary of the event-specific query lexicons used to construct the observed discourse corpora. The purpose of this figure is methodological rather than inferential: it shows how each crisis event was operationalized during data collection through hashtags, political actors, slogans, institutional references, locations, and event-specific keywords. For example, election-related events are anchored around partisan and campaign vocabulary, while protest, legal, public-health, and geopolitical events are characterized by more domain-specific terms such as movement slogans, court-related language, mandate terminology, or conflict actors. This helps make the scraping strategy more transparent and allows readers to assess whether the observed corpora were seeded with coherent event-level vocabularies. Importantly, these word clouds should not be interpreted as empirical term-frequency results from the collected posts; the empirical lexical analysis is reported separately through the TF--IDF divergence results. Instead, the figure documents the data-collection layer and complements the paired event design.

\subsection{Event Typology and Sociological Context}

To avoid treating events as interchangeable topical containers, we characterize them according to their sociological structure. Some events, such as elections and court decisions (as in Table~\ref{tab:event-typology}), are institutionally mediated and produce relatively formal public discourse. Others, such as protests, shootings, and geopolitical escalations, are faster-moving, decentralized, and more likely to generate fragmented or colloquial responses. This typology allows us to interpret the Caricature Gap not only as a model-level artifact, but also as an interaction between generative systems and the social structure of the event being simulated.

\begin{table}[t]
\centering
\caption{Sociological characterization of the nine crisis events. The typology is used as an interpretive layer for comparing event-level Caricature Gaps rather than as a causal classification.}
\label{tab:event-typology}
\small
\begin{tabularx}{\textwidth}{L{0.25\textwidth}L{0.19\textwidth}Y L{0.16\textwidth}}
\toprule
Event & Event type & Public discourse structure & Expected heterogeneity \\
\midrule
COVID-19 pandemic & Public-health crisis & Policy-driven, prolonged, and institutionally mediated & Medium \\
Jan 6 Capitol attack & Political rupture & Highly polarized, event-specific, and adversarial & High \\
2020 US election & Electoral event & Institutional, partisan, and campaign-centered & Medium--High \\
2024 US election & Electoral event & Partisan, candidate-centered, and identity-linked & High \\
Dobbs/Roe v. Wade & Legal-political event & Institutional, juridical, and movement-based & Medium \\
2020 BLM protests & Protest movement & Decentralized, grassroots, and locally situated & High \\
US midterms & Electoral event & Institutional, campaign-based, and comparatively routinized & Medium \\
Utah shooting & Local violent event & Fast-moving, localized, and identity-linked & High \\
US-Iran war & Geopolitical conflict & International, militarized, and escalation-oriented & High \\
\bottomrule
\end{tabularx}
\end{table}

\subsection{Event Pairing and Harmonization}

The key design principle of the corpus is event pairing. Rather than studying a single topic, we compare observed and synthetic discourse across multiple heterogeneous events. This reduces the risk that a result is driven by one topic alone and allows us to examine whether divergence is stable or event-specific. Here, paired refers to event alignment rather than one-to-one matching or equal sample size across observed and synthetic corpora.

All text was normalized before analysis. We removed URLs, HTML fragments, platform-specific artifacts, and obvious encoding anomalies. We retained the underlying lexical content needed for sentiment, toxicity, and lexical-framing analysis.

\subsection{Dataset Summary}

Table~\ref{tab:data_summary} reports, for each event, the number of observed and synthetic posts, the platforms represented, and the approximate date range.

\begin{table}[t]
\centering
\caption{Summary of the paired multi-event dataset. The observed corpus may include human-authored posts, bots, coordinated activity, and machine-assisted text; therefore, the comparison is between observed and synthetic discourse rather than verified human and verified AI authorship.}
\label{tab:data_summary}
\scriptsize
\setlength{\tabcolsep}{3pt}
\renewcommand{\arraystretch}{1.08}
\begin{tabularx}{\textwidth}{@{}Y r r L{0.22\textwidth}L{0.22\textwidth}@{}}
\toprule
Event & Observed & Synthetic & Platforms & Date range \\
\midrule
COVID-19 pandemic & 59{,}442 & 50{,}000 & Twitter & Jan 2020--Dec 2022 \\
Jan 6 Capitol attack & 279{,}617 & 20{,}000 & Twitter, Telegram & Jan 2021--Feb 2021 \\
2020 US election & 72{,}711 & 50{,}000 & Twitter, Telegram, Reddit & Jun 2020--Nov 2020 \\
2024 US election & 126{,}733 & 150{,}936 & Twitter, Telegram, Reddit & Jun 2024--Nov 2024 \\
Dobbs/Roe v. Wade & 11{,}065 & 30{,}000 & Twitter & May 2022--Aug 2022 \\
2020 BLM protests & 47{,}577 & 30{,}000 & Twitter & May 2020--Sep 2020 \\
US midterm elections & 31{,}227 & 30{,}000 & Twitter & Jan 2022--Nov 2022 \\
Utah shooting & 5{,}257 & 6{,}000 & Twitter & Sep 2025--Oct 2025 \\
US-Iran war & 588{,}841 & 200{,}000 & YouTube & Feb 2026--Mar 2026 \\
\midrule
Total & 1{,}222{,}470 & 566{,}936 & --- & --- \\
\bottomrule
\end{tabularx}
\end{table}

\subsection{Synthetic Generation Protocol}

Because the synthetic corpus is central to the interpretation of the audit, we document the generation procedure separately from the observed-data collection procedure. The goal of the synthetic corpus is not to imitate individual users, but to produce event-aligned political discourse populations that can be compared with observed online discourse at the aggregate level. Table~\ref{tab:generation-protocol} summarizes the protocol. In all cases, direct observed posts were not inserted into the generation prompt; the synthetic side was conditioned on event descriptions, topical framing, and, where applicable, political-role or stance cues (Cf. repository).

\begin{table}[t]
\centering
\caption{Synthetic discourse generation protocol. The synthetic corpus was generated using ChatGPT 4/5.5 between March 2025 and May 2026. Prompt templates and event-level generation logs should be released with the code to ensure reproducibility.}
\label{tab:generation-protocol}
\small
\begin{tabularx}{\textwidth}{L{0.28\textwidth}Y}
\toprule
Component & Description \\
\midrule
Model & ChatGPT 4/5.5. \\
Generation period & Prompts were executed between March 2025 and May 2026, aligned with the event-specific synthetic corpora. \\
Prompt type & Event-conditioned political-discourse prompts that describe the crisis event and request social-media-style posts. \\
Decoding parameters & ChatGPT interface default generation settings were used; exact temperature/top-$p$ values were not manually configured in the interface. \\
Observed posts in prompt & No direct observed posts were included in the generation prompt. \\
Persona/stance conditioning & None, political-role conditioning, or event-role conditioning depending on the event design. \\
Post-processing & Normalization, removal of malformed outputs, deduplication, and language filtering. \\
Release practice & Aggregate statistics and code are shared; potentially harmful synthetic examples are handled with caution. \\
\bottomrule
\end{tabularx}
\end{table}

\section{Measures and Analytical Framework}

We analyze divergence between observed and synthetic discourse using four complementary dimensions.

\subsection{Pillar 1: Emotional Intensity}

We estimate sentiment using VADER compound scores~\cite{hutto2014vader}. This provides a lightweight measure of polarity suitable for short social-media texts. We interpret these values comparatively rather than as direct psychological measurements.

To assess harmful affective tone, we additionally score toxicity using a transformer-based classifier (\texttt{unitary/toxic-bert}~\cite{detoxify2020}), which returns a continuous score on $[0,1]$. We treat this as a model-based indicator of hostile language rather than a direct measure of intent.

\subsection{Pillar 2: Structural Regularity}

To capture differences in discourse texture, we compute basic structural features including word count and punctuation ratio. These measures are not meant to capture stylistic quality, but population-level regularity. The key question is whether generated discourse exhibits narrower and more homogeneous distributions than observed discourse.

\subsection{Pillar 3: Lexical-Ideological Framing}

We estimate lexical divergence using TF-IDF-based ranking of unigrams and bigrams after removing structural noise. The goal is not topic modeling in the strict sense, but identification of salient lexical markers that differentiate observed and synthetic discourse within each event. We use these lexical profiles to assess whether synthetic discourse is more abstract, formalized, or ideologically compressed.

\subsection{Pillar 4: Cross-Event Dependency}

To summarize event-level divergence, we define a simple gap measure:
\[
\Delta = \left| \mu_{\text{syn}} - \mu_{\text{obs}} \right|,
\]
where $\mu_{\text{syn}}$ and $\mu_{\text{obs}}$ denote the synthetic and observed means for a given metric within an event. We refer to this quantity as the \textit{Caricature Gap}. Its purpose is descriptive: it allows us to compare the magnitude of divergence across events and across dimensions.

\subsection{Dispersion and Distributional Interpretation}

Although the Caricature Gap provides an interpretable event-level summary, mean differences alone cannot capture all forms of population distortion. We therefore interpret the mean gaps together with dispersion summaries, especially standard deviations and implied variance ratios $\sigma^2_{\mathrm{syn}}/\sigma^2_{\mathrm{obs}}$. These quantities are particularly important for the structural and sentiment analyses because the central claim concerns population compression: synthetic discourse may differ from observed discourse not only in average sentiment, toxicity, or length, but also in the breadth and shape of the corresponding distributions. In the reusable audit protocol, these summaries can be further extended with full distributional distances, such as Wasserstein distance or Jensen--Shannon divergence, when the goal is to compare different generators, prompts, or platform-specific subsets.

\subsection{Event-Moderation Lens}

To connect the statistical gaps to the sociological event typology, we use the event categories as an interpretive moderation lens rather than as causal ground truth. The guiding expectation is that gaps should be larger for fast-moving and decentralized events, where observed discourse is more heterogeneous, and smaller for institutionally mediated events, where public discourse is more formalized. Because the study contains nine events, we treat this analysis as descriptive and hypothesis-generating rather than as a fully powered event-level regression.

\subsection{Statistical Reporting}

For the quantitative metrics, we report descriptive statistics (mean and standard deviation for both corpora), the absolute Caricature Gap $\Delta$, and effect size measured by Cohen's $d$ (positive values indicate synthetic $>$ observed). We complement these with 95\% bootstrap confidence intervals on the mean difference ($\mu_{\text{syn}} - \mu_{\text{obs}}$, 2{,}000 resamples), and a two-sided Mann--Whitney $U$ test. Given the large corpus size, virtually all comparisons reach conventional significance thresholds; we therefore foreground effect size, confidence intervals, and dispersion patterns as the primary inferential quantities. The quantitative results are summarized in Table~\ref{tab:stats}.
\section{Results}
Table~\ref{tab:stats} summarizes the descriptive statistics, effect sizes, and bootstrap confidence intervals for sentiment, toxicity, and word count across all paired events; Mann--Whitney tests were used to assess rank-order differences. We discuss each pillar in turn below.

\begin{table}
\centering
\caption{Statistical summary for the three quantitative pillars. $\mu_{\text{obs}}$ and $\mu_{\text{syn}}$ are corpus means; $d$ is Cohen's $d$ (positive = synthetic $>$ observed); 95\% CI is bootstrapped on $\mu_{\text{syn}}-\mu_{\text{obs}}$ (2{,}000 resamples).}
\label{tab:stats}
\begin{tabular}{llrrrrrr}
\toprule
Metric & Event & $\mu_{\text{obs}}$ & $\sigma_{\text{obs}}$ & $\mu_{\text{syn}}$ & $\sigma_{\text{syn}}$ & $d$ & 95\% CI \\
\midrule
\multirow{10}{*}{Sentiment}
 & All events                      &  0.018 & 0.522 & $-$0.215 & 0.458 & $-$0.47 & [$-$0.246,\;$-$0.221] \\
 & 2020 BLM Protests               & $-$0.077 & 0.558 & $-$0.285 & 0.307 & $-$0.46 & [$-$0.239,\;$-$0.177] \\
 & 2020 US Election    &  0.012 & 0.527 & $-$0.360 & 0.440 & $-$0.77 & [$-$0.410,\;$-$0.315] \\
 & 2024 US Election                &  0.067 & 0.528 & $-$0.412 & 0.448 & $-$0.98 & [$-$0.514,\;$-$0.442] \\
 & COVID-19 Pandemic               &  0.078 & 0.519 &  0.015 & 0.398 & $-$0.14 & [$-$0.095,\;$-$0.029] \\
 & Dobbs / Roe v. Wade             &  0.053 & 0.517 &  0.154 & 0.360 &  0.22 & [0.069,\;0.134] \\
 & Jan 6 Capitol Attack            & $-$0.017 & 0.411 & $-$0.382 & 0.441 & $-$0.85 & [$-$0.395,\;$-$0.334] \\
 & US Midterm Elections            &  0.065 & 0.538 & $-$0.080 & 0.457 & $-$0.29 & [$-$0.179,\;$-$0.109] \\
 & US-Iran War         & $-$0.180 & 0.480 & $-$0.467 & 0.424 & $-$0.61 & [$-$0.320,\;$-$0.265] \\
 & Utah Shooting                   & $-$0.035 & 0.542 & $-$0.262 & 0.410 & $-$0.47 & [$-$0.263,\;$-$0.192] \\
\midrule
\multirow{10}{*}{Toxicity}
 & All events                      & 0.082 & 0.209 & 0.079 & 0.206 & $-$0.02 & [$-$0.008,\;0.002] \\
 & 2020 BLM Protests               & 0.154 & 0.288 & 0.002 & 0.005 & $-$0.75 & [$-$0.168,\;$-$0.138] \\
 & 2020 US Election    & 0.080 & 0.194 & 0.110 & 0.220 &  0.14 & [0.015,\;0.045] \\
 & 2024 US Election                & 0.082 & 0.209 & 0.313 & 0.341 &  0.82 & [0.211,\;0.252] \\
 & COVID-19 Pandemic               & 0.026 & 0.113 & 0.005 & 0.015 & $-$0.27 & [$-$0.028,\;$-$0.016] \\
 & Dobbs / Roe v. Wade             & 0.093 & 0.214 & 0.003 & 0.009 & $-$0.59 & [$-$0.101,\;$-$0.079] \\
 & Jan 6 Capitol Attack            & 0.095 & 0.216 & 0.188 & 0.276 &  0.38 & [0.077,\;0.111] \\
 & US Midterm Elections            & 0.032 & 0.139 & 0.069 & 0.190 &  0.22 & [0.025,\;0.049] \\
 & US-Iran War         & 0.140 & 0.260 & 0.038 & 0.121 & $-$0.53 & [$-$0.115,\;$-$0.090] \\
 & Utah Shooting                   & 0.098 & 0.224 & 0.014 & 0.076 & $-$0.50 & [$-$0.096,\;$-$0.072] \\
\midrule
\multirow{10}{*}{Word Count}
 & All events                      & 32.16 & 55.93 & 23.08 & 11.29 & $-$0.23 & [$-$10.16,\;$-$8.11] \\
 & 2020 BLM Protests               & 25.90 & 13.55 & 21.97 &  7.54 & $-$0.36 & [$-$4.77,\;$-$3.16] \\
 & 2020 US Election    & 38.22 & 68.53 & 22.50 &  9.80 & $-$0.27 & [$-$18.90,\;$-$12.50] \\
 & 2024 US Election                & 35.04 & 65.42 & 21.78 &  9.59 & $-$0.28 & [$-$16.81,\;$-$9.89] \\
 & COVID-19 Pandemic               & 28.16 & 11.95 & 20.64 &  7.26 & $-$0.76 & [$-$8.31,\;$-$6.83] \\
 & Dobbs / Roe v. Wade             & 26.20 & 12.94 & 26.83 &  9.63 &  0.06 & [$-$0.24,\;1.42] \\
 & Jan 6 Capitol Attack            & 23.40 & 52.35 & 22.29 &  3.79 & $-$0.03 & [$-$3.72,\;1.44] \\
 & US Midterm Elections            & 54.75 & 100.29 & 23.36 & 10.91 & $-$0.44 & [$-$36.80,\;$-$26.51] \\
 & US-Iran War         & 52.00 & 78.30 & 31.39 & 11.06 &  -0.34 & [-23.10,\; -16.10] 

 \\
 & Utah Shooting                   & 25.63 & 44.70 & 16.34 & 18.19 & $-$0.27 & [$-$12.32,\;$-$7.09] \\
\bottomrule
\end{tabular}
\end{table}

\subsection{Pillar 1: Emotional Intensity}

\begin{figure}
    \centering
    \includegraphics[width=0.5\textwidth]{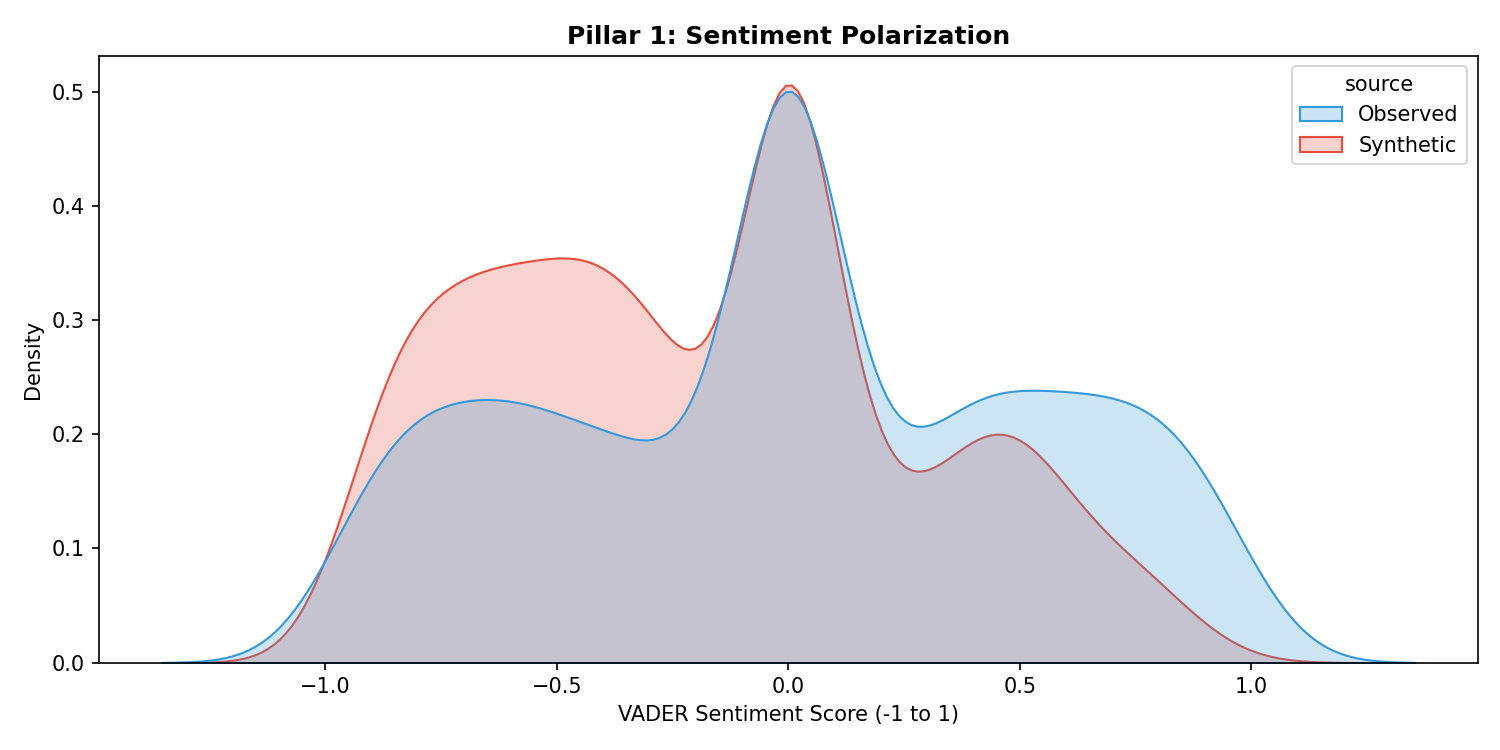}
    \caption{Aggregate sentiment polarization: observed discourse remains near-neutral while synthetic discourse anchors strongly in the negative register.}
    \label{fig:sentiment_agg}
\end{figure}

\begin{figure}
    \centering
    \includegraphics[width=0.9\textwidth]{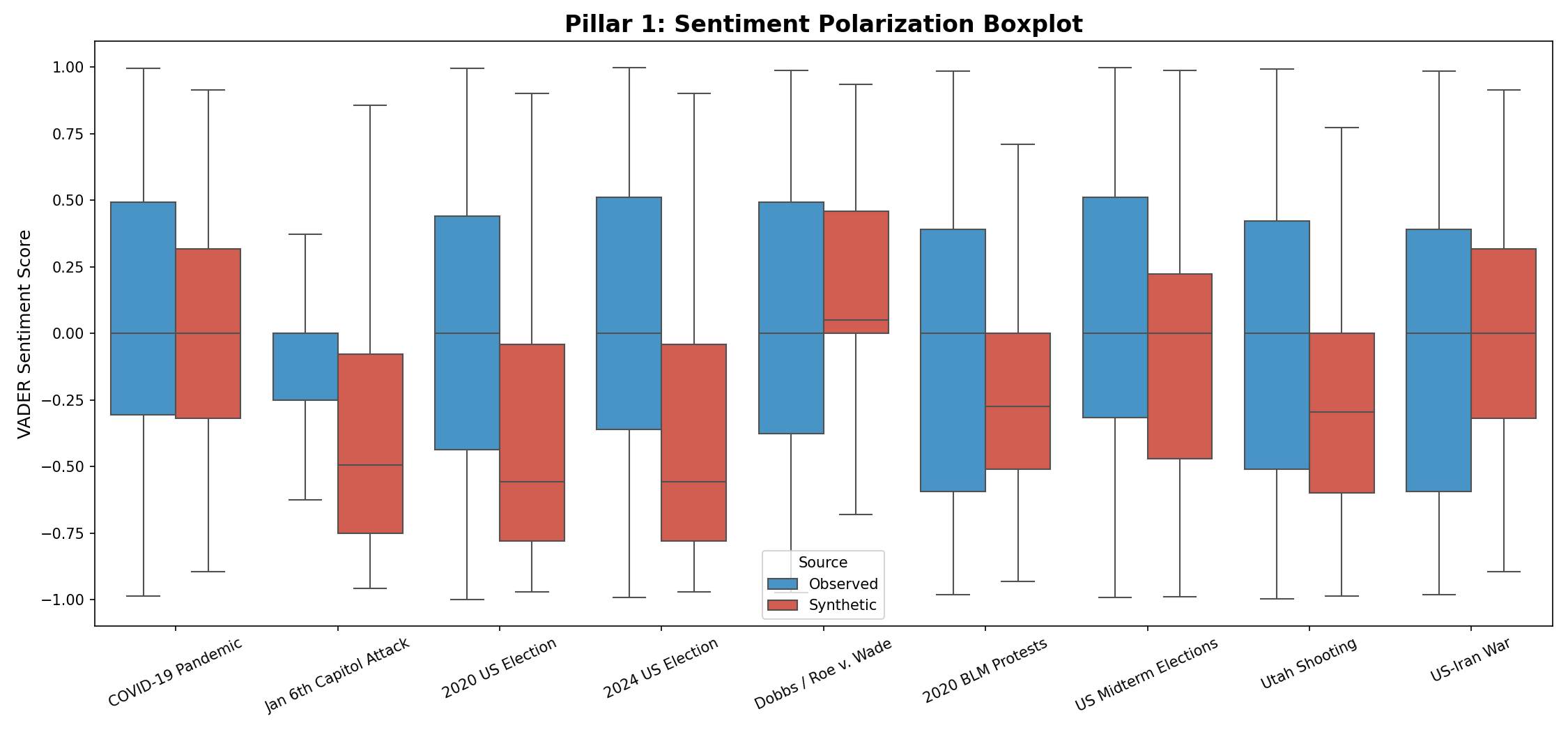}
    \caption{Per-event whisker bars for sentiment polarization variance. The tight clustering of synthetic scores (red) versus the wide spread of observed scores (blue) is consistent across all nine events.}
    \label{fig:sentiment_boxplot}
\end{figure}

Across the full corpus, observed discourse remains close to neutral on average ($\mu_{\text{obs}} = +0.018$, $\sigma = 0.52$), whereas synthetic discourse is substantially more negative ($\mu_{\text{syn}} = -0.215$, $\sigma = 0.46$; $d = -0.47$, 95\% CI $[-0.246, -0.221]$, $p < .001$). The 2024 US Election shows the largest single-event gap ($d = -0.98$), while COVID-19 shows the smallest ($d = -0.14$), consistent with the event-dependency hypothesis. Notably, Dobbs/Roe v. Wade is the only event where synthetic discourse is \emph{more positive} than observed ($d = +0.22$), likely reflecting the formal, legal register of the prompts used to generate that corpus. Figure~\ref{fig:sentiment_agg} shows the aggregate sentiment density, and Figure~\ref{fig:sentiment_boxplot} presents per-event whisker bars that quantify the variance and interquartile range of this polarization.

\begin{figure}
    \centering
    \includegraphics[width=0.7\textwidth]{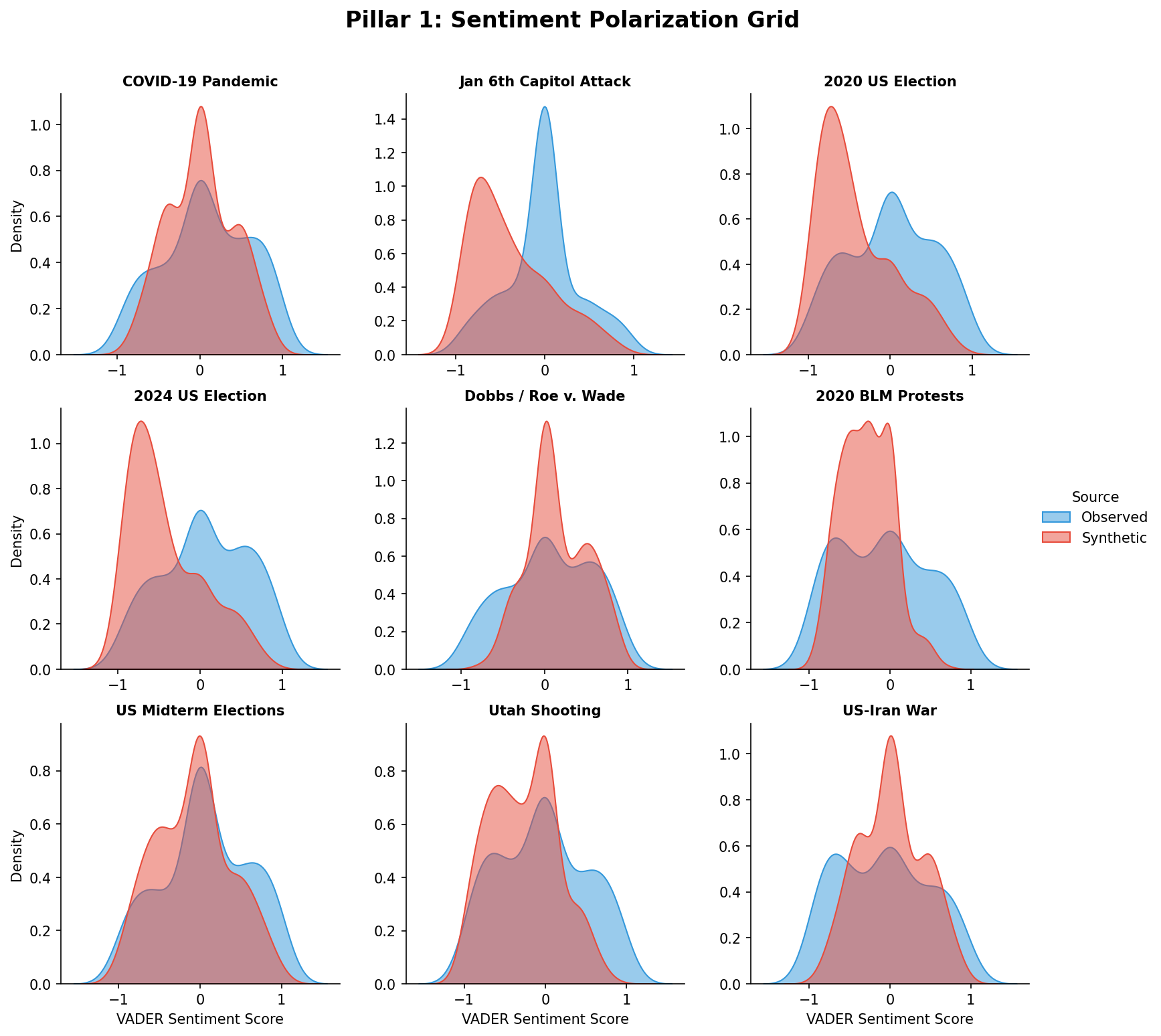}
    \caption{Faceted sentiment grids across all nine events.}
    \label{fig:sentiment_faceted}
\end{figure}

\begin{figure}
    \centering
    \includegraphics[width=0.5\textwidth]{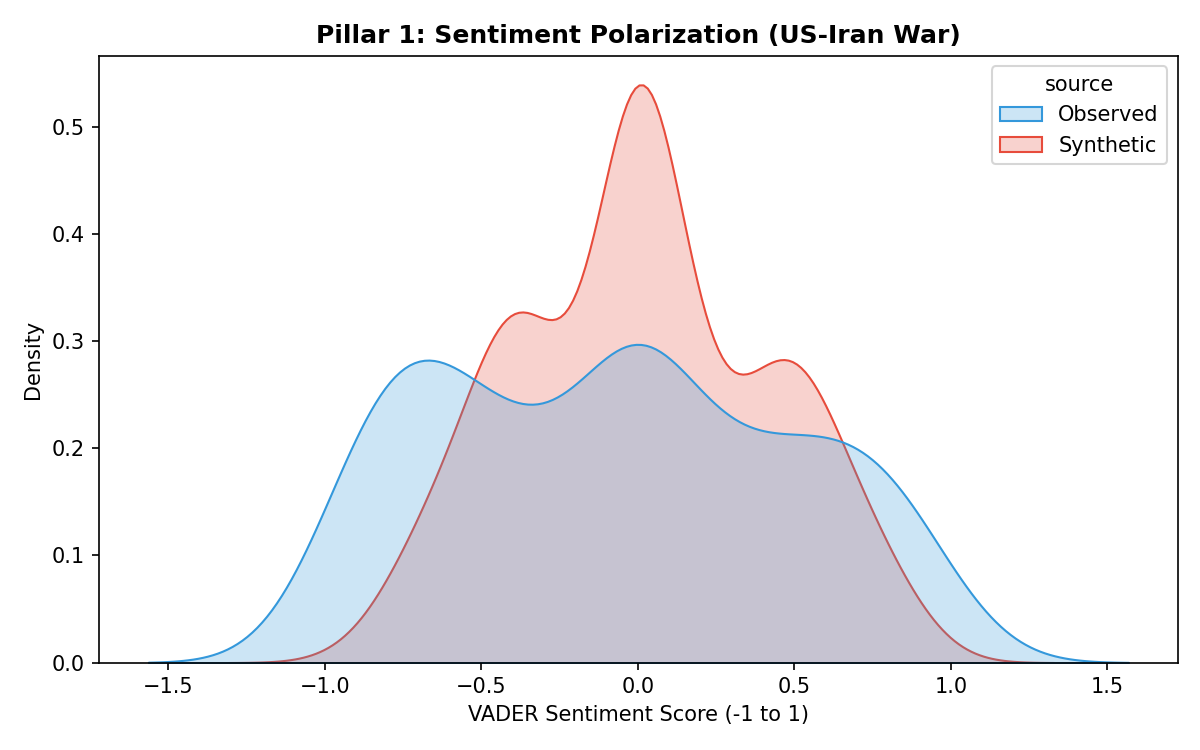}
    \caption{Isolated sentiment polarization for the US-Iran War context.}
    \label{fig:sentiment_us_iran}
\end{figure}

When the full nine-event grid is faceted (Figure~\ref{fig:sentiment_faceted}), local emotional topography is exposed---each event exhibits its own observed discourse signature that synthetic discourse fails to reproduce. The US-Iran War (Figure~\ref{fig:sentiment_us_iran}) is highlighted as a standalone case given the unique geopolitical register of the discourse.

\begin{figure}
    \centering
    \includegraphics[width=0.6\textwidth]{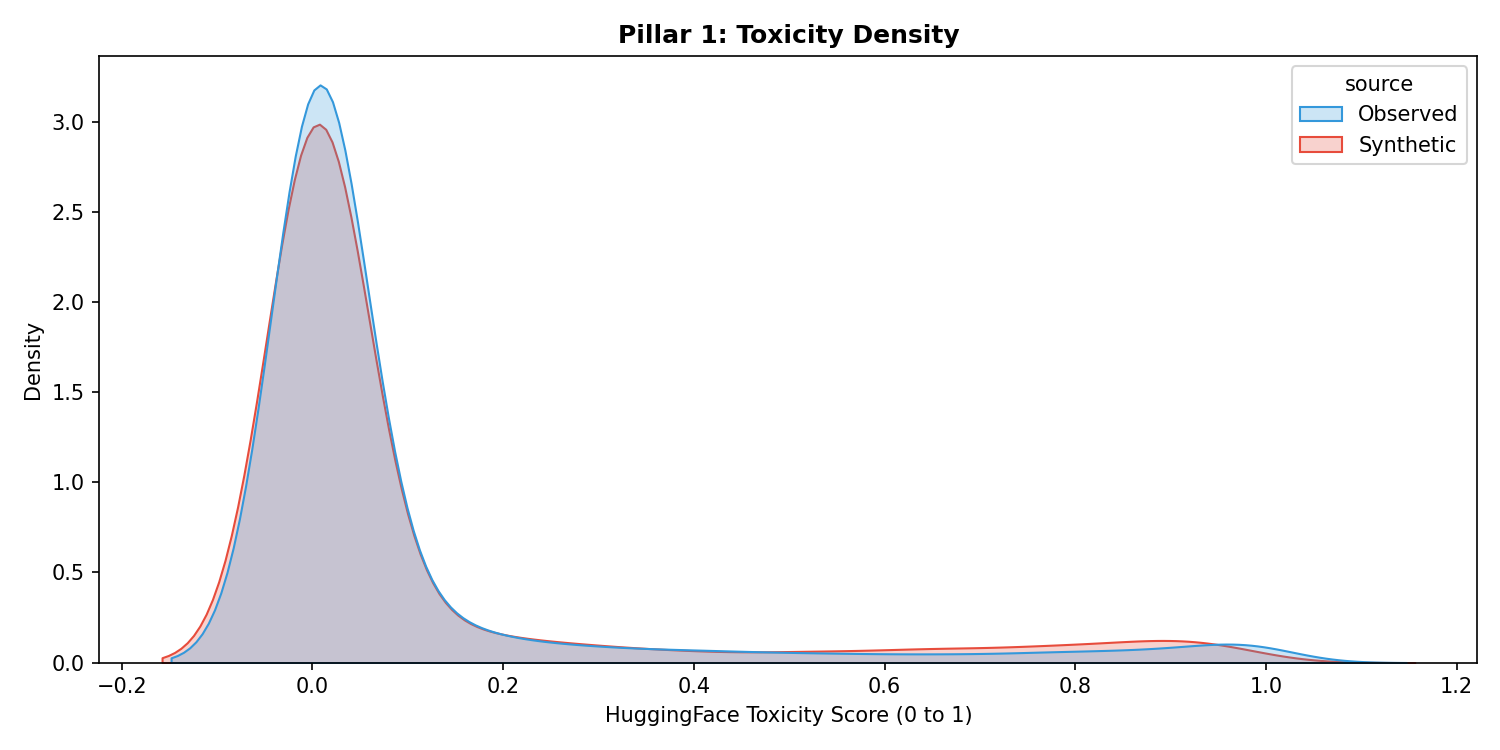}
    \caption{Aggregate toxicity density: global toxicity bounds are nearly identical across both sources.}
    \label{fig:toxicity_agg}
\end{figure}

\begin{figure}
    \centering
    \includegraphics[width=0.7\textwidth]{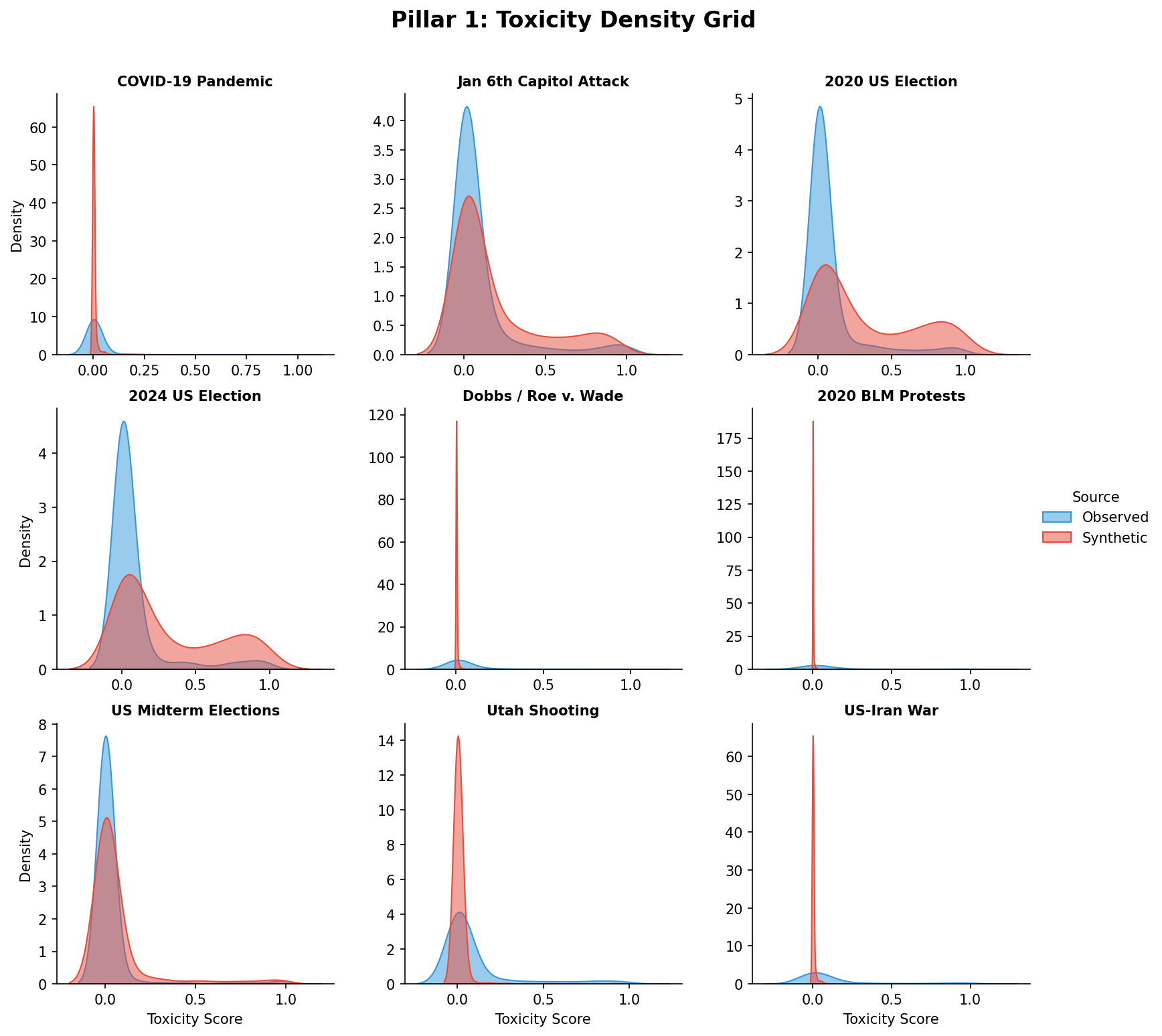}
    \caption{Faceted toxicity splits by event. The generator attempts to match organic harassment distributions (blue), but exhibits patterns consistent with safety-constrained generation (red).}
    \label{fig:toxicity_faceted}
\end{figure}

By contrast, aggregate toxicity levels are much closer between the two sources of discourse. Figure~\ref{fig:toxicity_agg} confirms that average toxicity occupies similar global ranges in both populations, even when sentiment differs more clearly, and the faceted view (Figure~\ref{fig:toxicity_faceted}) reinforces this finding at the per-event level.

This combination---more negative sentiment without proportionally higher toxicity---suggests that synthetic discourse may be better described as affectively intensified than overtly more toxic.

\subsection{Pillar 2: Structural Regularity}

\begin{figure}
    \centering
    \includegraphics[width=1.0\textwidth]{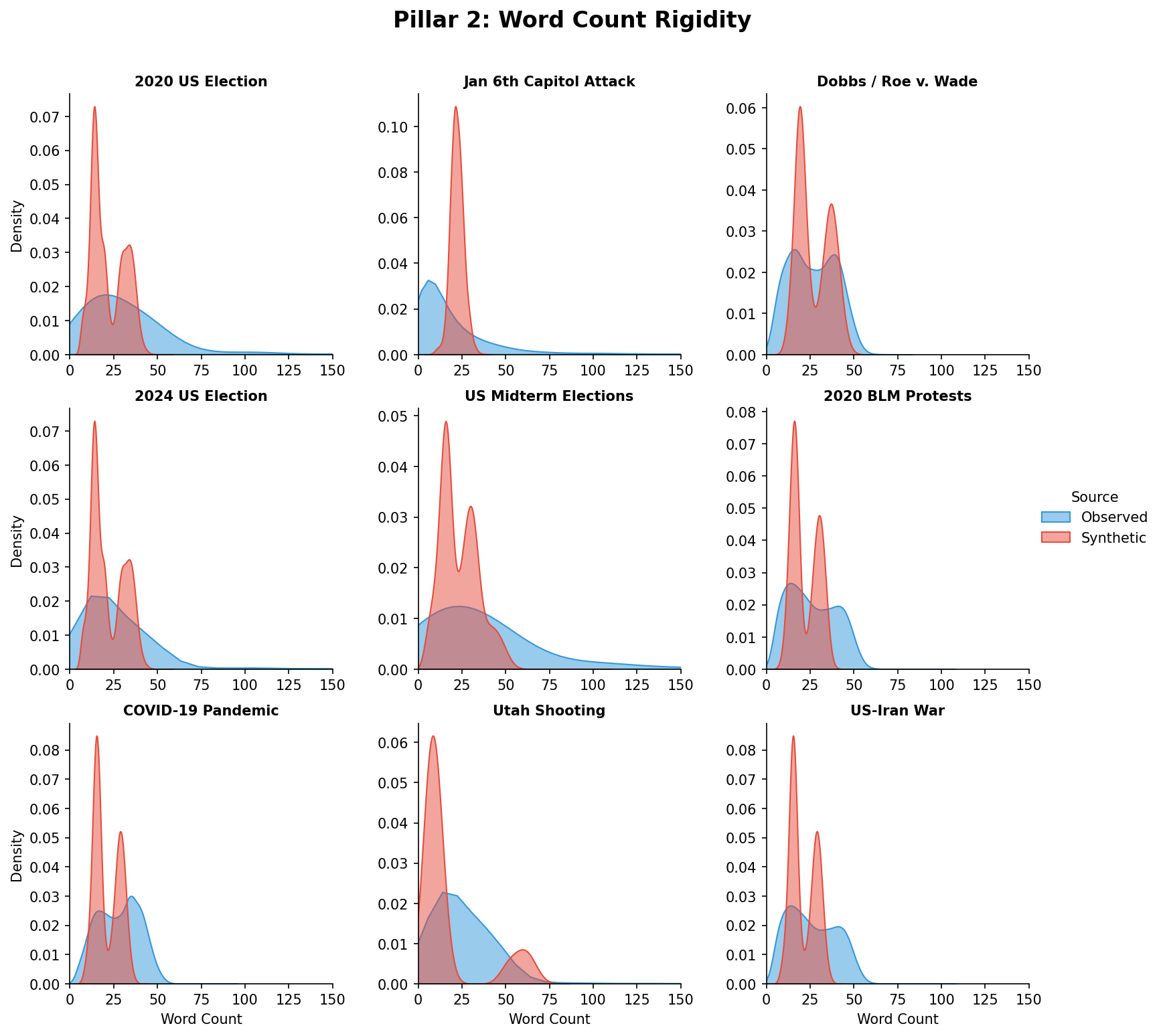}
    \caption{Faceted word-count distributions across nine events. Synthetic discourse (red) produces an identical structural spike regardless of event context, while observed discourse (blue) adapts organically.}
    \label{fig:wordcnt_faceted}
\end{figure}

Observed discourse exhibits broad, long-tailed structural distributions, particularly in word count. Across the corpus, observed posts average 32.2 words ($\sigma = 55.9$), while synthetic posts average 23.1 words ($\sigma = 11.3$; $d = -0.23$, 95\% CI $[-10.16, -8.11]$, $p < .001$). The key signal is not the mean difference but the variance: the synthetic standard deviation is five times smaller than observed, confirming population-level structural homogenization. The effect is strongest for COVID-19 ($d = -0.76$) and negligible for Dobbs/Roe v.\ Wade ($d = 0.06$, CI crosses zero) and Jan~6 ($d = -0.03$)---events whose formal, institutional register happens to align with the model's default output style. Synthetic discourse, in contrast, is more concentrated around narrower length bands. Figure~\ref{fig:wordcnt_faceted} illustrates this pattern across all nine events.

\begin{figure}
    \centering
    \includegraphics[width=0.85\textwidth]{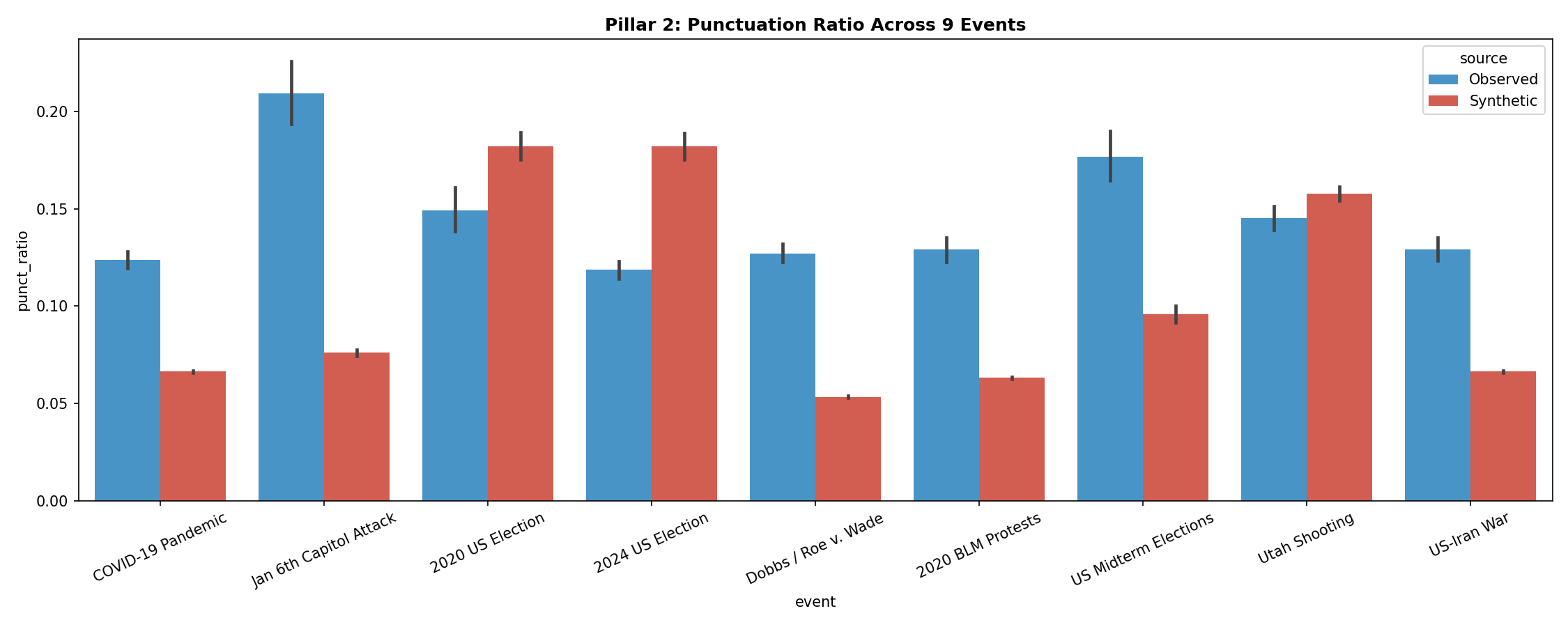}
    \caption{Punctuation ratio variance across events.}
    \label{fig:punct}
\end{figure}

A similar difference appears in punctuation use. Figure~\ref{fig:punct} shows that synthetic discourse is more regular in sentence completion patterns, whereas observed discourse is structurally noisier and more variable.

Together, these findings indicate that synthetic discourse exhibits \textit{regressive length homogenization}: regardless of the sociological context of the event being simulated, the model collapses toward its median token-generation threshold.

The distributional interpretation strengthens this conclusion. A mean-only comparison understates the magnitude of structural collapse because observed posts contain long tails, short fragments, replies, slogans, and platform-specific artifacts, whereas synthetic posts concentrate around a narrower length band. Median, IQR, and variance-ratio summaries therefore provide a more direct measure of the population-level compression that motivates the term \textit{algorithmic caricature}.

\subsection{Pillar 3: Lexical-Ideological Framing}

\begin{figure}
    \centering
    \includegraphics[width=0.85\textwidth]{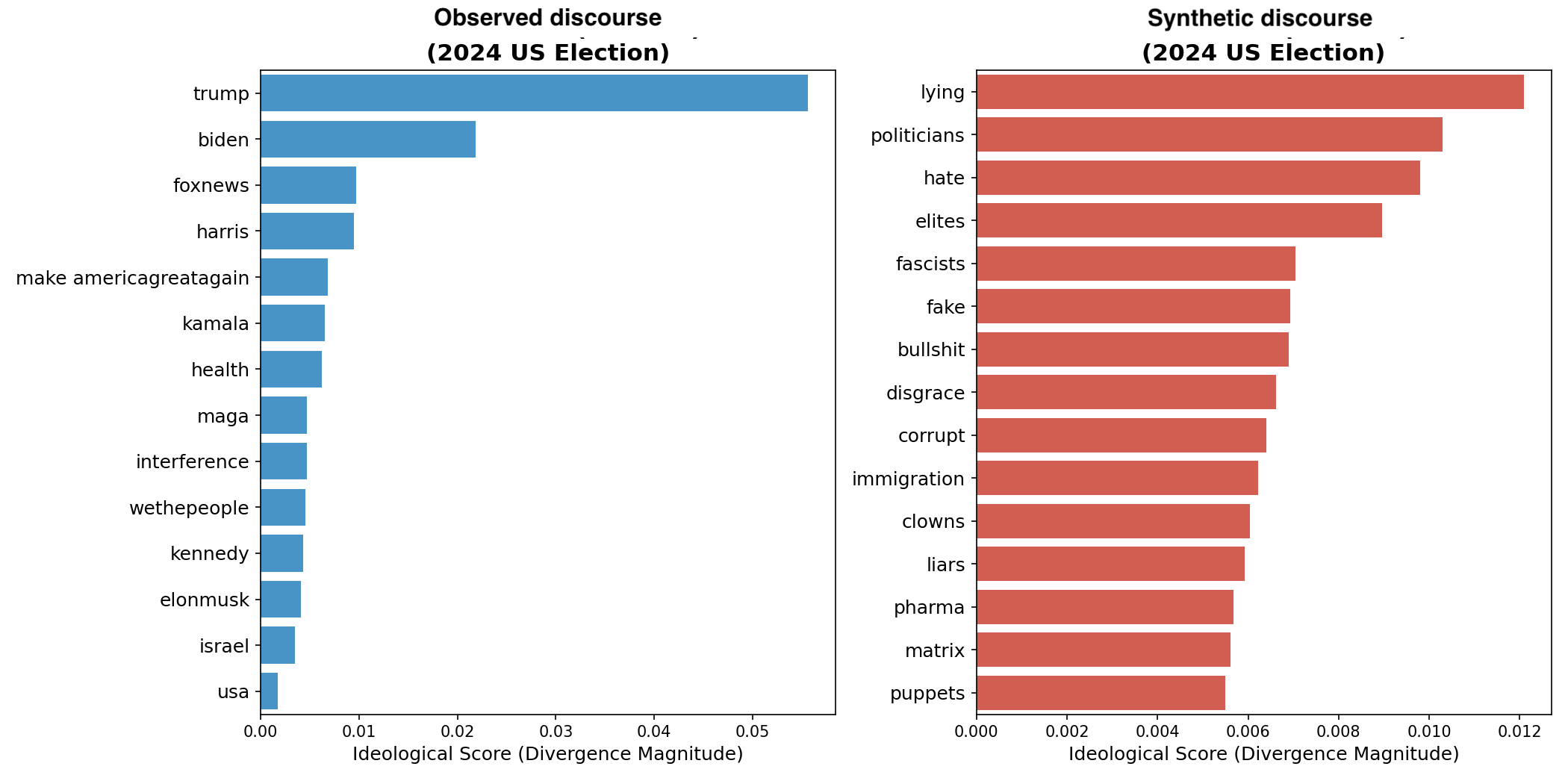}
    \caption{Lexical divergence during the 2024 US election.}
    \label{fig:tfidf_1}
\end{figure}

\begin{figure}
    \centering
    \includegraphics[width=0.85\textwidth]{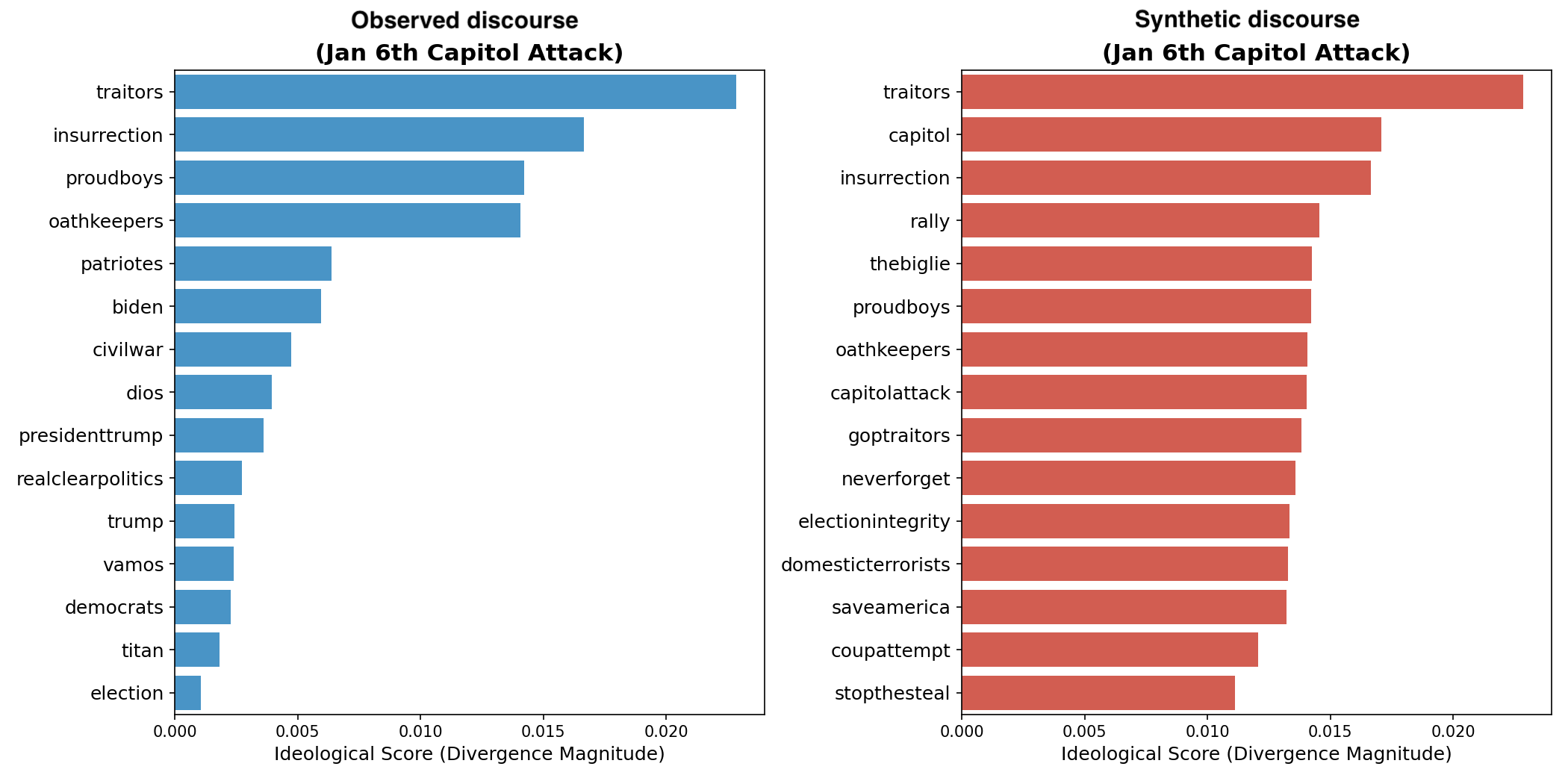}
    \caption{Lexical divergence surrounding the Jan 6 Capitol attack.}
    \label{fig:tfidf_2}
\end{figure}

\begin{figure}
    \centering
    \includegraphics[width=0.85\textwidth]{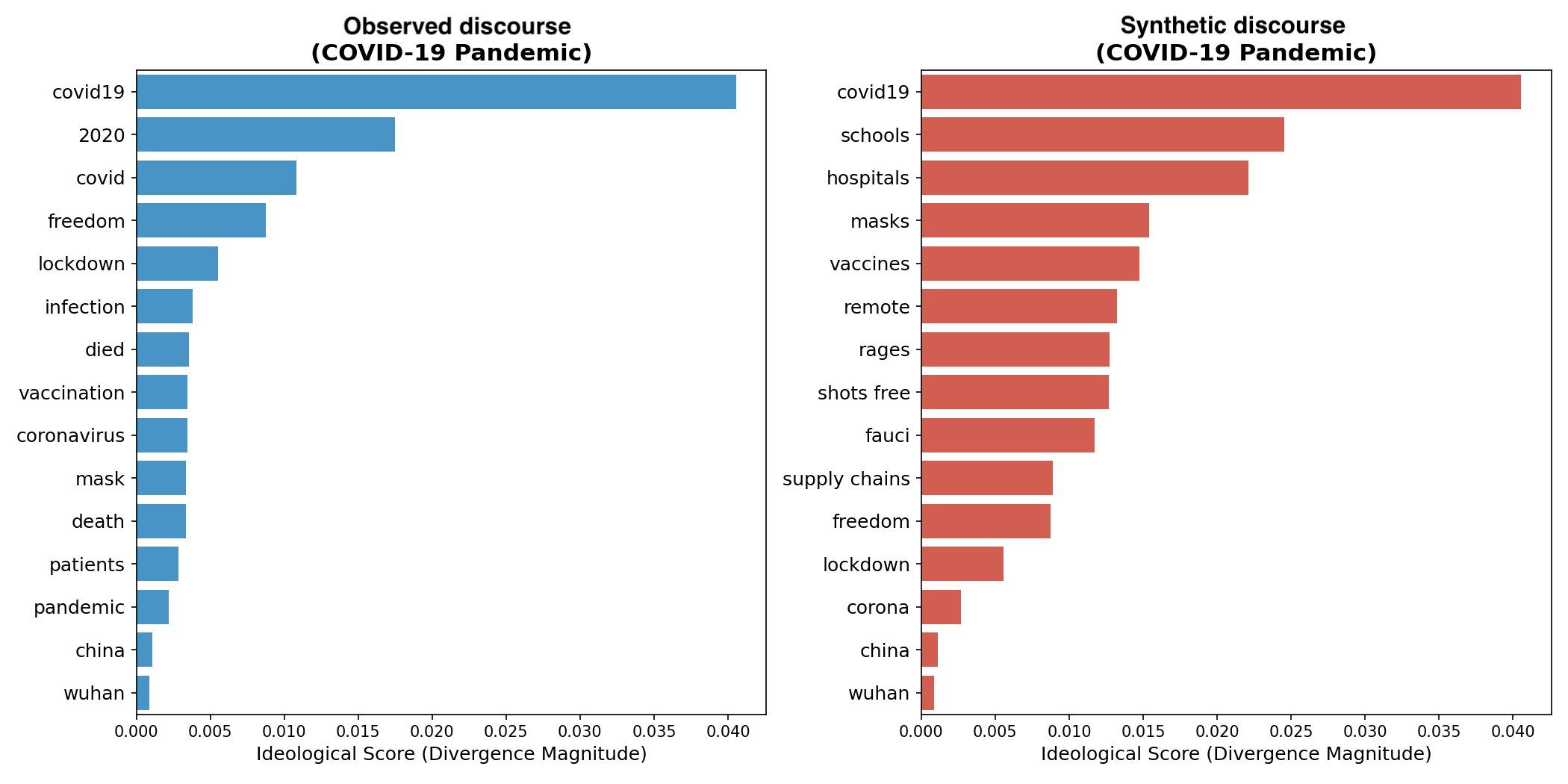}
    \caption{Lexical divergence concerning the COVID-19 pandemic.}
    \label{fig:tfidf_3}
\end{figure}

\begin{figure}
    \centering
    \includegraphics[width=0.85\textwidth]{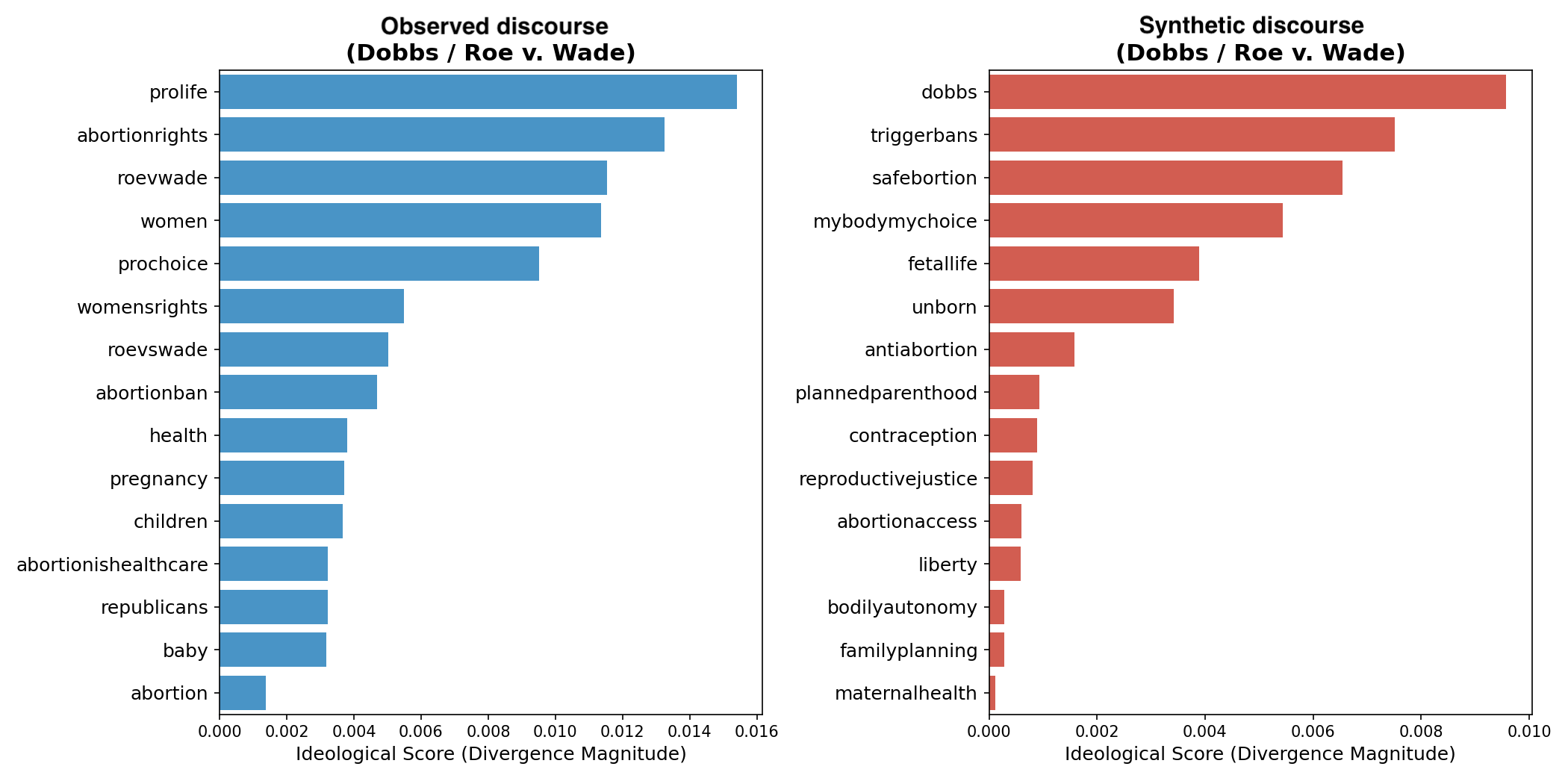}
    \caption{Lexical divergence concerning Dobbs/Roe v. Wade.}
    \label{fig:tfidf_4}
\end{figure}

\begin{figure}
    \centering
    \includegraphics[width=0.85\textwidth]{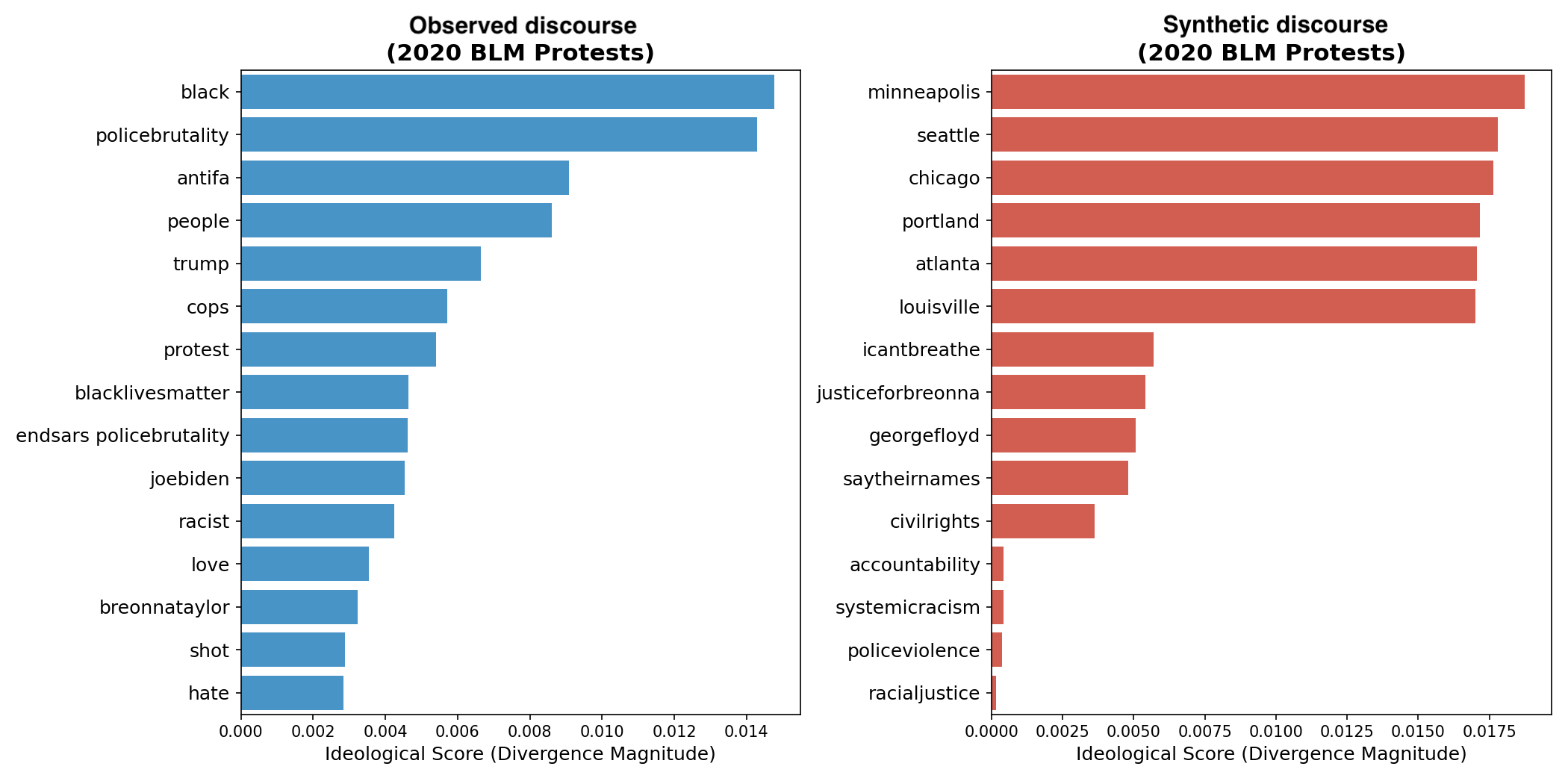}
    \caption{Lexical divergence concerning the 2020 BLM protests.}
    \label{fig:tfidf_5}
\end{figure}

\begin{figure}
    \centering
    \includegraphics[width=0.85\textwidth]{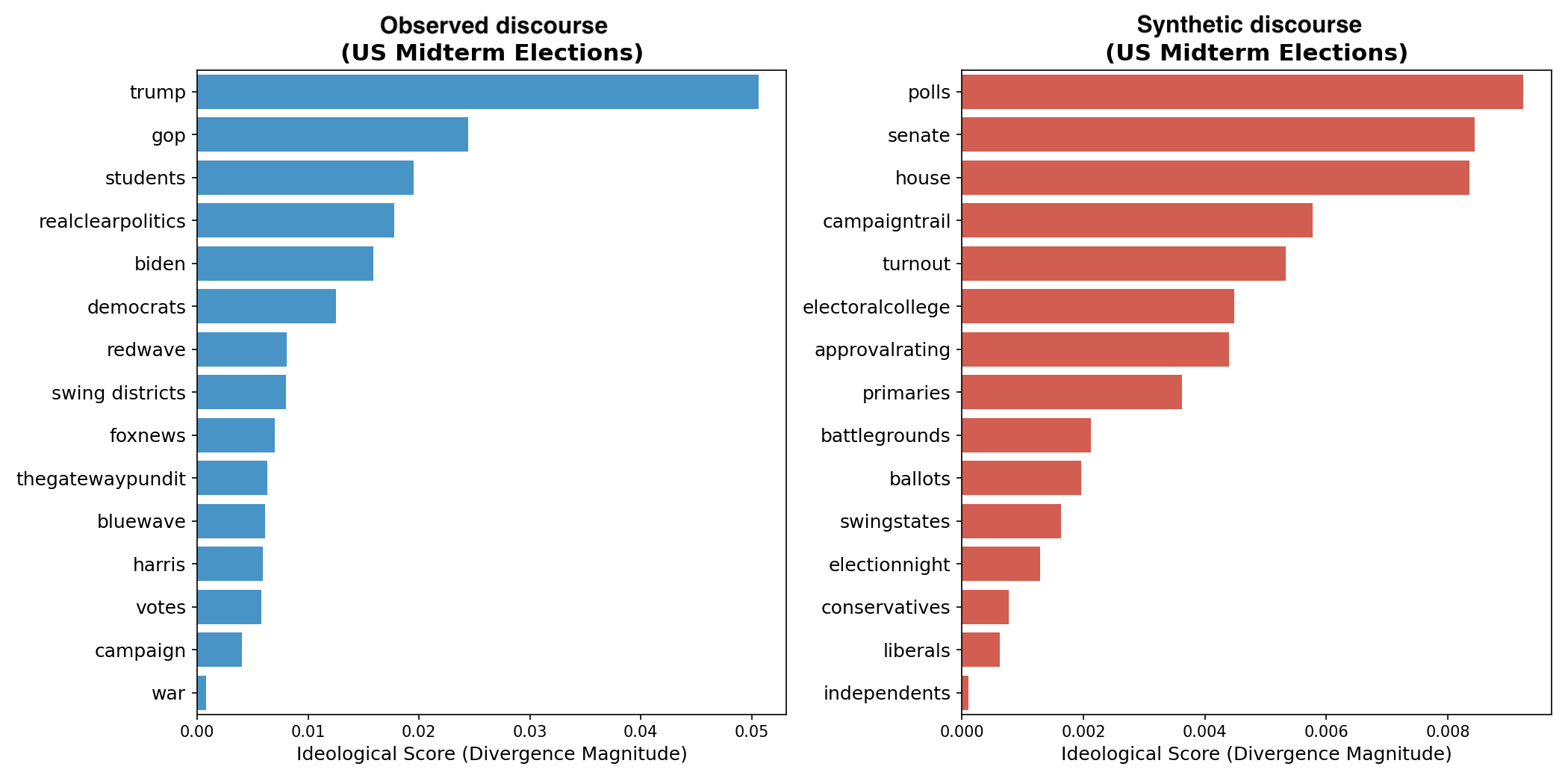}
    \caption{Lexical divergence concerning the US midterm elections.}
    \label{fig:tfidf_6}
\end{figure}

\begin{figure}
    \centering
    \includegraphics[width=0.85\textwidth]{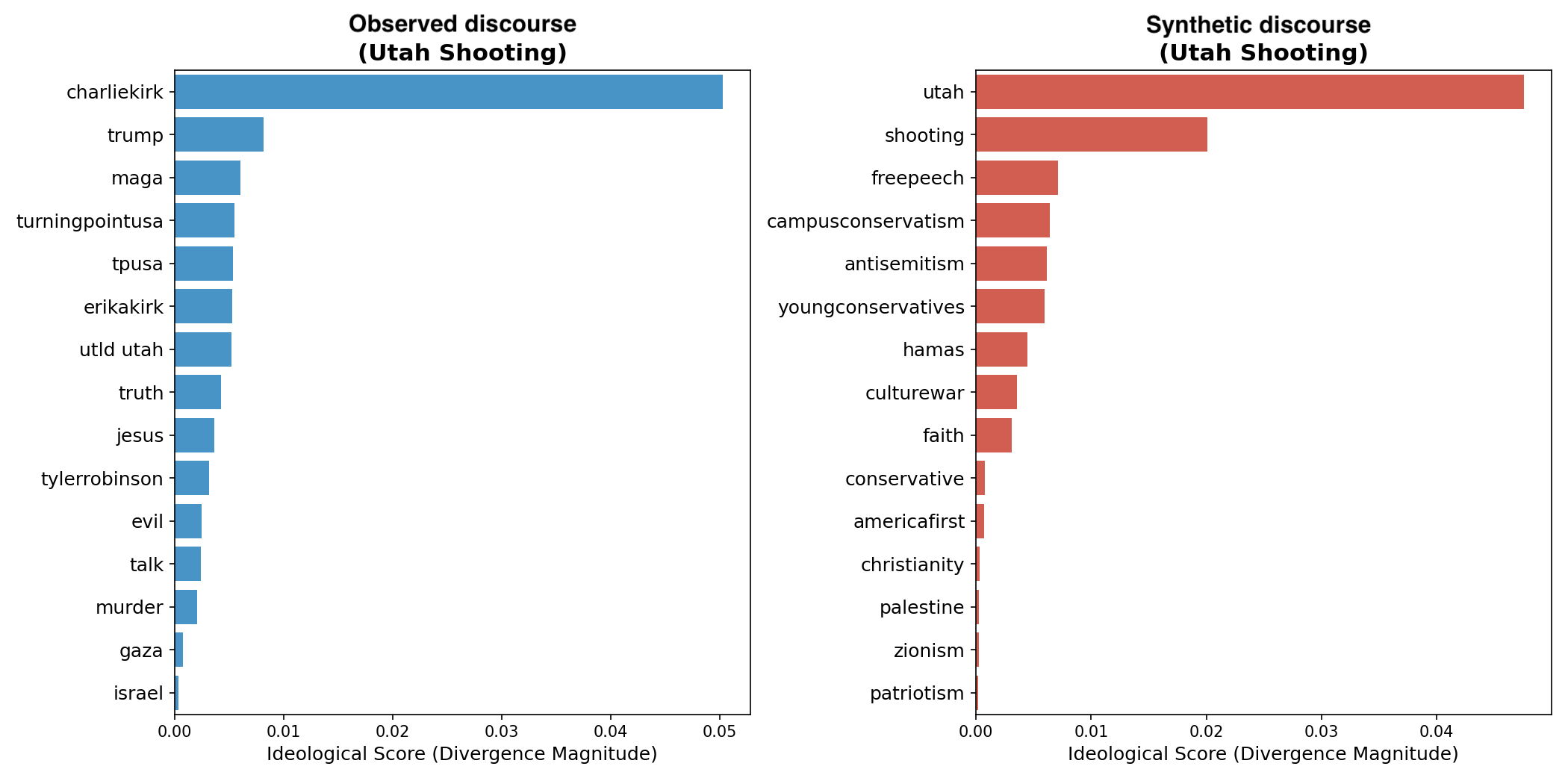}
    \caption{Lexical divergence concerning the Utah shooting.}
    \label{fig:tfidf_7}
\end{figure}

\begin{figure}
    \centering
    \includegraphics[width=0.85\textwidth]{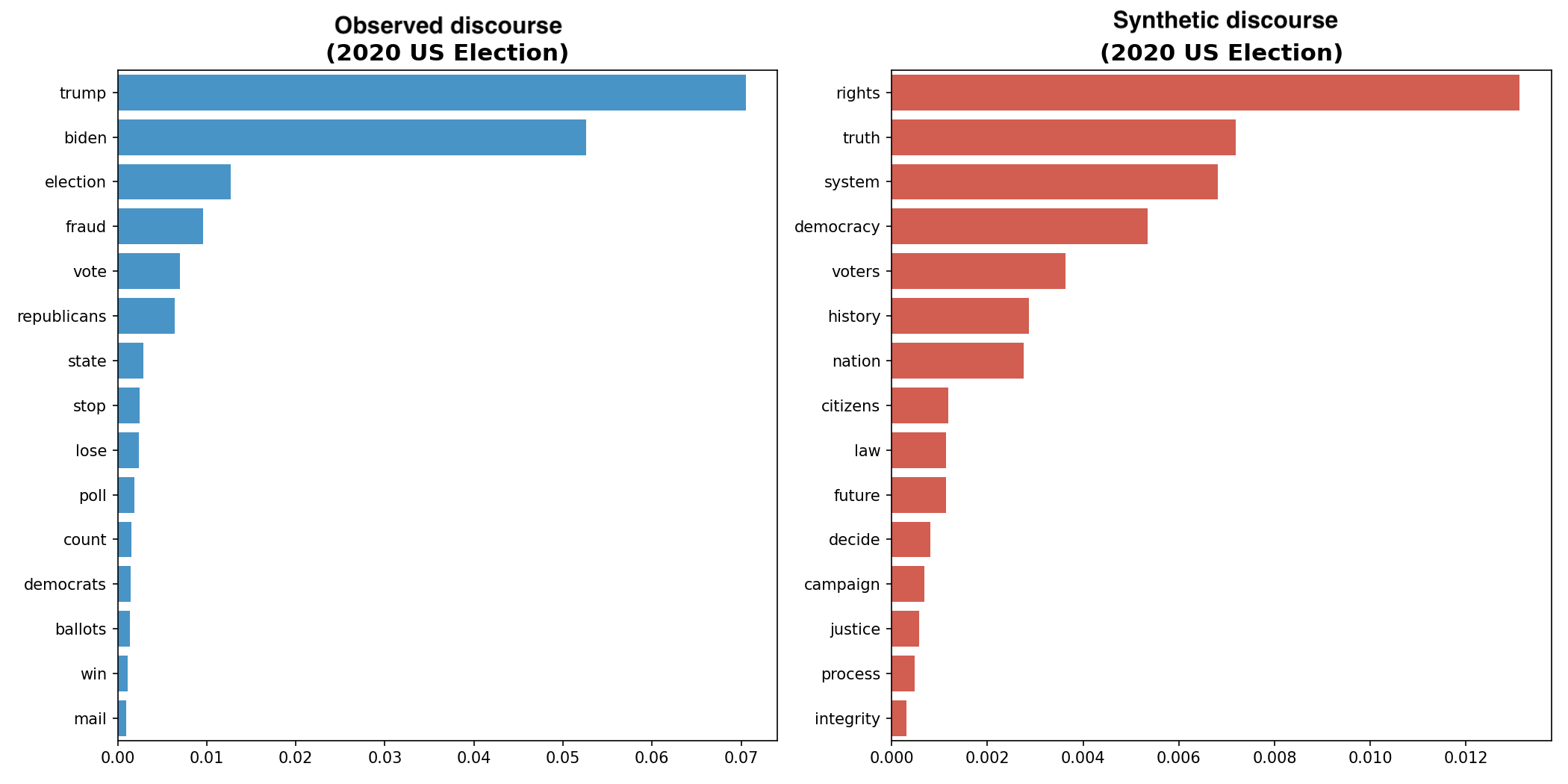}
    \caption{Lexical divergence isolating the 2020 US election paradigm.}
    \label{fig:tfidf_8}
\end{figure}

\begin{figure}
    \centering
    \includegraphics[width=0.85\textwidth]{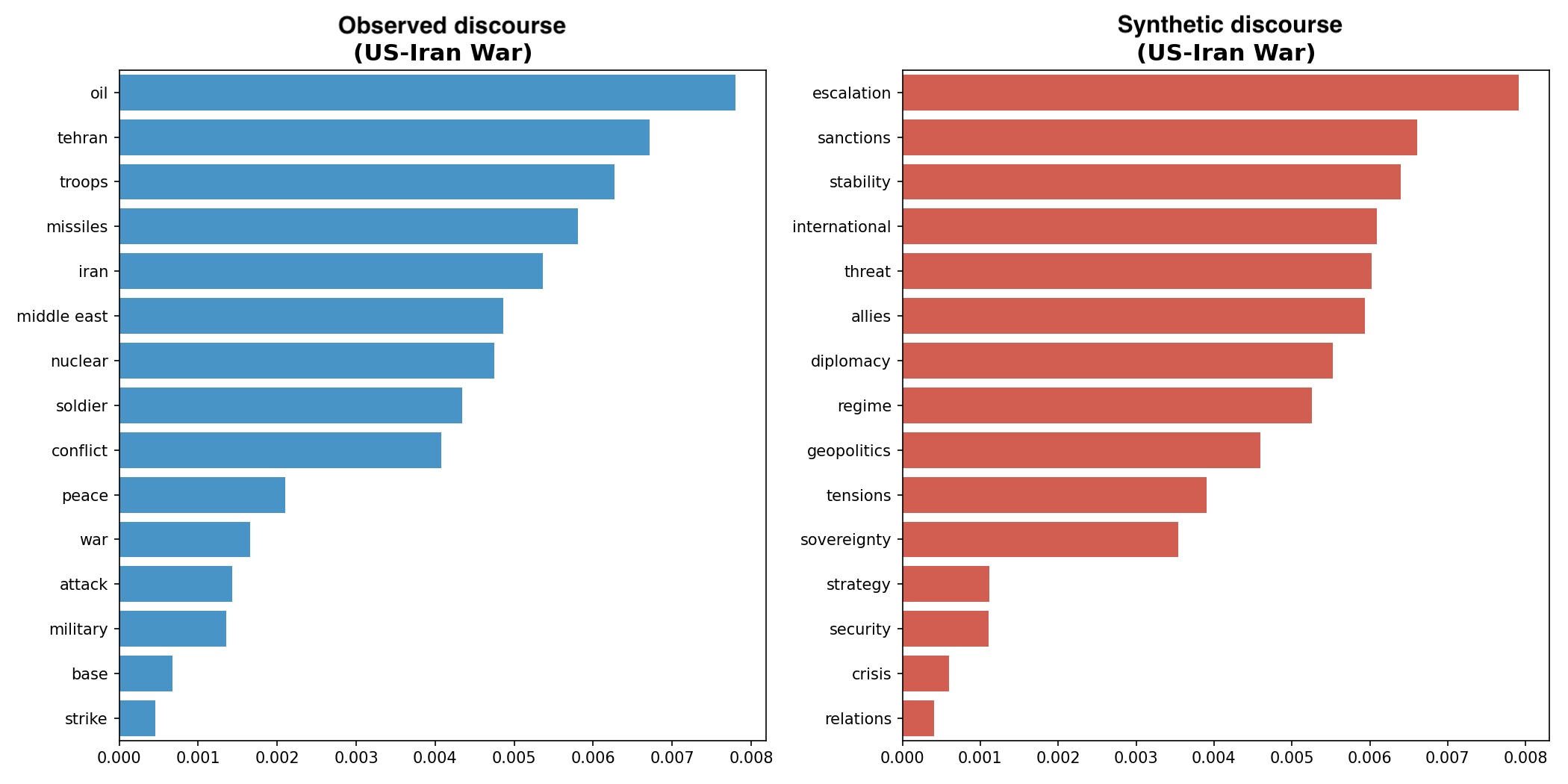}
    \caption{Lexical divergence evaluating the geopolitical US-Iran War.}
    \label{fig:tfidf_9}
\end{figure}

The lexical profiles of observed and synthetic discourse differ substantially across all nine events. Observed discourse foregrounds concrete event-specific markers, named actors, and colloquial references. Synthetic discourse more often emphasizes abstract or dramatized ideological language. Figures~\ref{fig:tfidf_1}--\ref{fig:tfidf_9} illustrate this pattern across events.

Overall, the lexical analysis confirms that synthetic discourse tends to compress ideological conflict into formalized or stereotyped rhetorical frames, whereas observed discourse remains closer to concrete, socially situated references.

\subsection{Pillar 4: Cross-Event Dependency}

\begin{figure}
    \centering
    \includegraphics[width=1.0\textwidth]{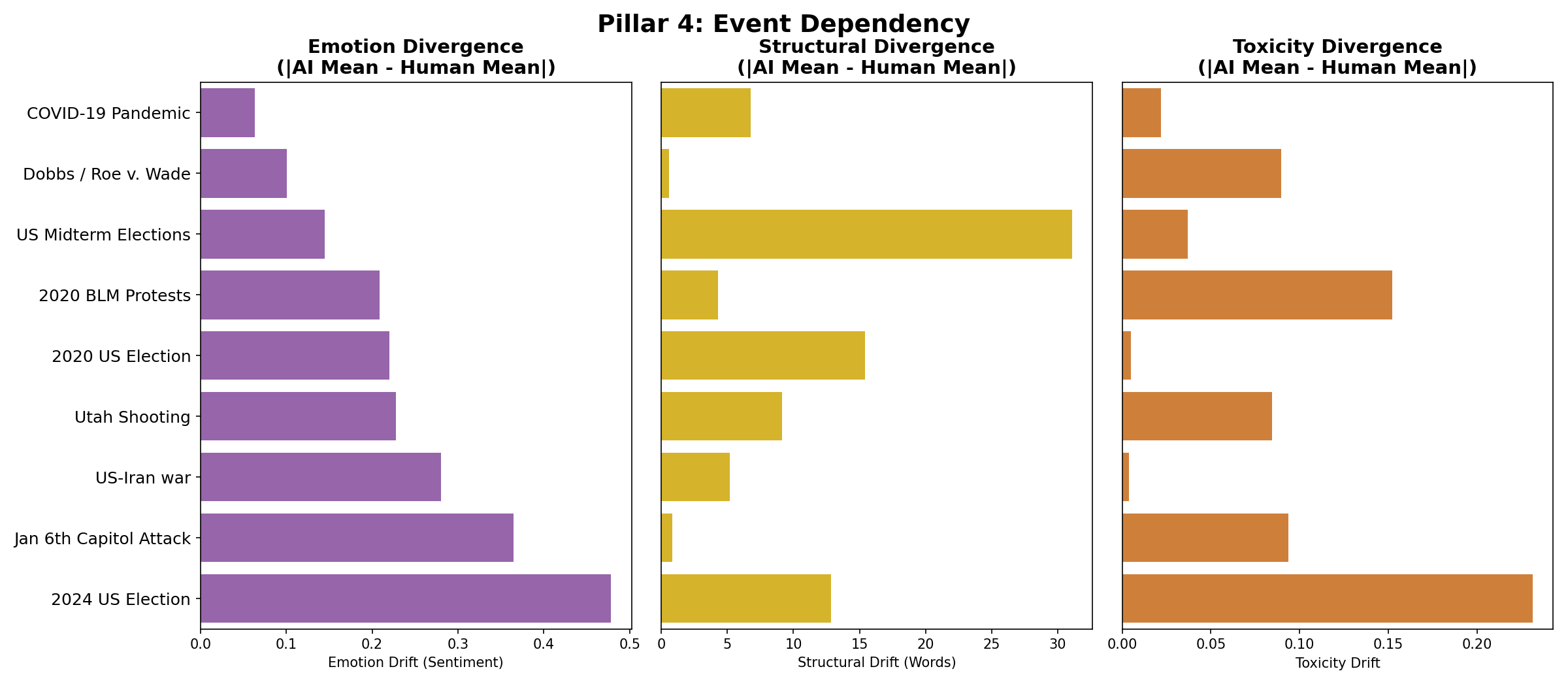}
    \caption{9-event cross-event Caricature Gap ($\Delta$) across sentiment, structural, and toxicity dimensions.}
    \label{fig:divergence}
\end{figure}

The divergence between observed and synthetic discourse is not constant across events. Figure~\ref{fig:divergence} summarizes the Caricature Gap ($\Delta = | \mu_{\text{syn}} - \mu_{\text{obs}} |$) across all nine events and three dimensions simultaneously.

For sentiment, the gap ranges from $\Delta = 0.063$ (COVID-19, $d = -0.14$) to $\Delta = 0.478$ (2024 US Election, $d = -0.98$), confirming that event type is a strong moderator of synthetic--observed divergence. The structural gap follows a similar gradient: word-count divergence is largest for the US Midterm Elections ($d = -0.44$) and essentially absent for Dobbs and Jan~6 ($d \approx 0$).

A particularly striking finding concerns toxicity. The direction of the gap \emph{reverses} across events: synthetic discourse is \emph{more} toxic than observed for the 2024 US Election ($d = +0.82$) and Jan~6 ($d = +0.38$), yet \emph{less} toxic for the BLM protests ($d = -0.75$), Utah ($d = -0.50$), and Dobbs ($d = -0.59$). This reversal suggests that the model's safety constraints interact with each event's dominant rhetorical register in non-uniform ways: the model may over-index on adversarial political language for election contexts, but under-produces the raw grassroots hostility characteristic of protest and legal-controversy discourse. A consistent pattern is that structural divergence is larger for rapid, decentralized events such as protest dynamics or election-period discourse, and smaller for events with more institutionalized or formalized public language. This suggests that the social realism of generated discourse depends not only on the model, but also on the sociological structure of the event being simulated.

Taken together, these patterns support an event-moderation interpretation: the Caricature Gap appears to vary with the sociological structure of the event. We treat this as descriptive rather than causal evidence. The typology in Table~\ref{tab:event-typology} is therefore used to interpret the observed pattern, not to identify a causal effect of event type on model behavior.

\section{Robustness and Sensitivity Considerations}
We conducted three lightweight robustness checks on the sampled enriched dataset. First, because the robustness subset contains balanced cohorts of 1,500 observed and 1,500 synthetic posts per event, equal-size sampling leaves the sentiment and toxicity gaps unchanged, confirming that the reported Caricature Gap is not driven by sample-size imbalance. Second, after removing duplicate text entries, the signed sentiment and toxicity gaps changed only negligibly, typically by less than 0.005. This indicates that the observed divergences are not artifacts of repeated synthetic generations, hashtag spam, or duplicated platform content. Third, we compared mean-based structural differences with median and interquartile-range summaries. Observed discourse has a median length of 23 words and an IQR of 26 words, whereas synthetic discourse has a median length of 21 words and an IQR of 15 words. Thus, the synthetic word-count IQR is approximately 42\% lower than the observed IQR, supporting the interpretation that LLM-generated discourse collapses toward a narrower structural range.

\begin{table}[t]
\centering
\caption{Robustness checks on the sampled enriched dataset. Sentiment and toxicity gaps are signed differences $(\mu_{\mathrm{syn}}-\mu_{\mathrm{obs}})$.}
\label{tab:robustness-results}
\scriptsize
\setlength{\tabcolsep}{4pt}
\renewcommand{\arraystretch}{1.08}
\begin{tabular}{lrrrr}
\toprule
Event & Sent. baseline & Sent. dedup. & Tox. baseline & Tox. dedup. \\
\midrule
COVID-19 pandemic & -0.0630 & -0.0631 & -0.0218 & -0.0219 \\
Jan 6 Capitol attack & -0.3644 & -0.3694 & 0.0936 & 0.0918 \\
2024 US election & -0.4784 & -0.4791 & 0.2314 & 0.2312 \\
Dobbs/Roe v. Wade & 0.1003 & 0.0942 & -0.0897 & -0.0916 \\
2020 BLM protests & -0.2083 & -0.2076 & -0.1522 & -0.1523 \\
Utah shooting & -0.2278 & -0.2260 & -0.0843 & -0.0848 \\
\midrule
Word count median/IQR & \multicolumn{2}{c}{Observed: 23.0 / 26.0} &
\multicolumn{2}{c}{Synthetic: 21.0 / 15.0} \\
\bottomrule
\end{tabular}
\end{table}

For these checks, the inferential emphasis should be on whether synthetic discourse remains more homogeneous and less event-specific than observed discourse. Because the dataset is very large, minor numeric differences are expected to be statistically significant; robustness should therefore be evaluated by the stability of the effect direction, the relative ranking of event-level gaps, and the persistence of dispersion-based compression.

\section{Discussion}

Our results point to a population-level limitation of synthetic political discourse. The main issue is not that generated text is grammatically poor. On the contrary, much of it is fluent. Rather, the issue is that it tends to underrepresent the heterogeneity of observed online populations. Compared with observed discourse, synthetic discourse is more affectively concentrated, structurally more regular, and lexically more abstract.

Two findings deserve particular attention. First, Dobbs/Roe v.\ Wade is the only event where synthetic sentiment is \emph{more positive} than observed ($d = +0.22$). One possible explanation is the formal, legal register of the prompts used to generate that corpus: the model's training on institutional text aligns naturally with juridical language, causing it to produce more measured, less emotionally charged output than the actual public reaction to the ruling. This illustrates a general principle---the model's behavioral bias is not fixed; it interacts with the register implied by the generation prompt.

Second, the toxicity reversal across events (synthetic more toxic for elections, less toxic for protests and legal controversy) challenges simple narratives about LLM safety constraints. Rather than uniformly suppressing hostility, the model appears to over-weight adversarial political framing in some contexts while under-producing grassroots register in others. This context-dependence is itself a diagnostic signal: it suggests that toxicity-based detection methods may need event-specific calibration rather than global thresholds.

This matters for two reasons. First, it suggests that population-level auditing may provide a useful complement to traditional AI-text detection. Even if local linguistic signatures weaken, generated discourse may still remain distinguishable through its aggregate behavioral profile. Second, it raises a substantive social-science question: if synthetic discourse increasingly participates in public debate, it may reshape political communication not only by scale, but by narrowing the forms of expression that circulate---privileging formal, dramatized language over the messy, colloquial texture of authentic digital mobilization.

The paper also has direct implications for evaluation practice. A model may appear highly realistic when judged on individual fluency, yet still fail to reproduce the emotional and structural diversity of a real discourse population. We therefore argue for a CSS-based auditing perspective that treats the \emph{population} rather than the \emph{post} as the unit of evaluation.

\section{Limitations and Ethical Considerations}

This study has several limitations. First, the observed corpus is not guaranteed to contain only human-authored content; it may include bots, coordinated campaigns, or machine-assisted posts. Our comparison is therefore between observed and synthetic discourse, not between verified human and verified AI text. Second, the synthetic corpus depends on the prompts, model choices, and generation procedures used to produce it. Different prompts, decoding settings, safety filters, or persona assumptions may change the magnitude of the observed gaps. Third, the Utah shooting and US-Iran war corpora may be less directly comparable to the other events because they differ in scale, recency, platform composition, and geopolitical context. They should therefore be interpreted as useful stress cases rather than as directly equivalent to the longer-running US domestic events.

Fourth, our measures are proxies. VADER does not capture the full nuance of political affect, toxicity classifiers are model-dependent, and TF--IDF lexical divergence is descriptive rather than causal. Finally, our interpretations of why these patterns emerge---for example, the possible role of safety constraints, training-data composition, or prompt register---should be treated as hypotheses consistent with the evidence rather than directly identified mechanisms.

The study also raises ethical and dual-use considerations. It involves politically sensitive discourse and synthetic political text that could be misused if redistributed without context. We therefore treat the dataset as a research artifact for aggregate analysis rather than as content to be reused for persuasion, imitation, or amplification. We do not attempt to identify individual users, infer private attributes, or attribute intent to specific accounts. Where possible, release should prioritize aggregate statistics, feature representations, reproducible code, and documented prompts while limiting the redistribution of harmful generated examples or raw politically sensitive content.

\section{A Reusable Audit Protocol}

Beyond the specific events analyzed here, the proposed framework can be used as a reusable audit protocol for future generative systems and simulated social-media environments. The protocol consists of eight steps: define the crisis event and collection window; construct a transparent seed lexicon; collect observed discourse under documented platform constraints; generate an aligned synthetic corpus under documented model and prompt settings; normalize both corpora using the same pipeline; compute the four divergence pillars; summarize gaps using mean-based and dispersion-aware measures; and interpret the results through an event typology and robustness checks. This makes the framework useful not only for retrospective analysis, but also for auditing new models, prompts, safety settings, or simulated online publics.

\section{Conclusion}

We presented a Computational Social Science framework for auditing synthetic political discourse across crisis events. Using a paired corpus spanning nine events, we showed that synthetic discourse differs systematically from observed discourse along four dimensions: emotional intensity, structural regularity, lexical-ideological framing, and cross-event dependency.

These findings suggest that the key limitation of synthetic discourse is not sentence-level fluency but reduced population realism. The proposed Caricature Gap and dispersion-aware audit protocol therefore shift evaluation from the isolated post to the discourse population. We argue that future work on synthetic-text auditing should move beyond local grammatical cues and ask whether generated discourse behaves like the online publics it claims to represent.

\bibliographystyle{apalike}
\bibliography{references}

\end{document}